\documentclass[lettersize,journal]{IEEEtran}
\usepackage{amsmath,amsfonts}
\usepackage{algorithmic}
\usepackage{algorithm}
\usepackage{array}
\usepackage[caption=false,font=normalsize,labelfont=sf,textfont=sf]{subfig}
\usepackage{textcomp}
\usepackage{stfloats}
\usepackage{url}
\usepackage{verbatim}
\usepackage{graphicx}
\usepackage{cite}
\usepackage[numbers,sort&compress]{natbib}
\usepackage{amsmath}
\usepackage{array}
\usepackage{color}
\usepackage{booktabs}
\usepackage{bm}
\usepackage{amssymb}
\usepackage{fancyhdr}
\usepackage[colorlinks,
            linkcolor=blue,
            anchorcolor=blue,
            citecolor=blue]{hyperref}          
\begin{document}

\title{Self-Supervised Monocular Depth Estimation \\ with Self-Reference Distillation and Disparity Offset Refinement}

\author{Zhong Liu, Ran Li*, Shuwei Shao, Xingming Wu and Weihai Chen  

\thanks{This paper was supported by the National Natural Science Foundation of
China under grant 61620106012.}
\thanks{Zhong Liu, Ran Li, Shuwei Shao, Xingming Wu and Weihai Chen are with the School of Automation
Science and Electrical Engineering, Beihang University, Beijing, China. (email: liuzhong@buaa.edu.cn, rnlee1998@buaa.edu.cn, swshao@buaa.edu.cn, wxmbuaa@163.com, whchen@buaa.edu.cn). }
\thanks{* Corresponding author: Ran Li}}



\maketitle

\thispagestyle{fancy}
\lfoot{Copyright \copyright 2023 IEEE. Personal use of this material is permitted. However, permission to use this material for any other purposes must be obtained from the IEEE by sending an email to pubs-permissions@ieee.org.}
\cfoot{}

\renewcommand{\headrulewidth}{0mm}
\begin{abstract}
Monocular depth estimation plays a fundamental role in computer vision. Due to the costly acquisition of depth ground truth, self-supervised methods that leverage adjacent frames to establish a supervision signal have emerged as the most promising paradigms. In this work, we propose two novel ideas to improve self-supervised monocular depth estimation: 1) self-reference distillation and 2) disparity offset refinement. Specifically, we use a parameter-optimized model as the teacher updated as the training epochs to provide additional supervision during the training process. The teacher model has the same structure as the student model, with weights inherited from the historical student model. In addition, a multiview check is introduced to filter out the outliers produced by the teacher model. Furthermore, we leverage the contextual consistency between high-level and low-level features to obtain multiscale disparity offsets, which are used to refine the disparity output incrementally by aligning disparity information at different scales. The experimental results on the KITTI and Make3D datasets show that our method outperforms previous state-of-the-art competitors.
\end{abstract}

\begin{IEEEkeywords}
Monocular depth estimation, Self-supervised learning, Self-reference distillation, Disparity alignment, Multiview check.
\end{IEEEkeywords}

\section{Introduction}
Perception of the 3D world is one of the main tasks in
computer vision. However, in many cases it might be unfeasible to obtain access to depth information relying on expensive or complex sensors. Depth estimation from a single image has gained extensive attention and has been shown to be a practical technology with applications ranging from localization, navigation, autonomous driving, and robot grasping to 3D reconstruction. 

In recent years, supervised monocular depth estimation has been widely studied and has made significant strides \cite{huang2022depth,song2021monocular,adabins,newcrfs,meng2021cornet}. Supervised depth estimation is a mapping problem from pixel-level RGB information to depth. With the aid of CNN, self-attention and other mechanisms, depth estimation is performed based on image texture, color information, and surrounding image relationships. While supervised depth estimation has achieved excellent performance, RGB-D data is still constrained in abundance and variety when compared with available RGB image and video data in the field. Furthermore, gathering a large number of accurate ground-truth datasets is a challenging task due to sensor noise and limited operating capabilities. Recent studies have identified a feasible alternative for performing depth estimator training in a self-supervised manner. The self-supervised method converts the depth estimation into an image synthesis using an intermediary variable (depth or disparity). For instance, the classical method Monodepth2 \cite{mono2} trains the model to predict the appearance of a target image from the viewpoint of another image, by minimizing the photometric reconstruction loss.

\begin{figure}[!t]
\centering
\includegraphics[width=0.46\textwidth]{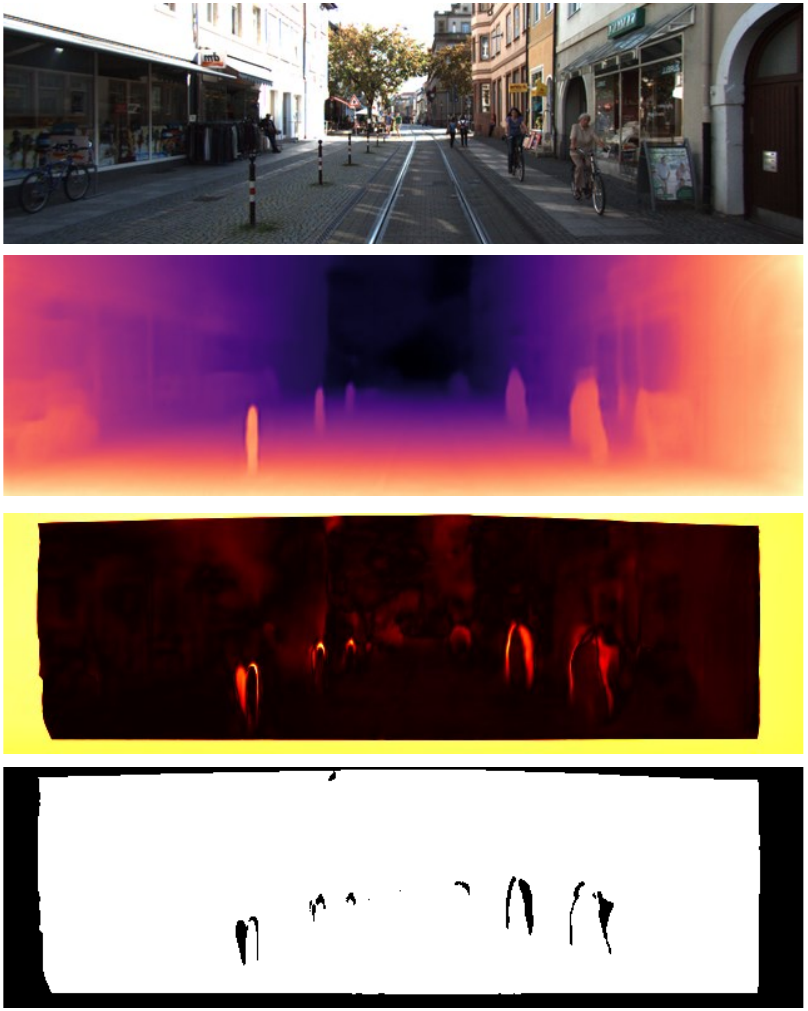}
\caption{Illustration of the depth map, error map and binary mask from the proposed method. The first row is the RGB image, and the second row is the depth map. The third row is the intermediate error map generated by the multiview check filter, where the places with large errors are red and yellowish. The fourth row is a binarized hard mask to filter out outliers.}
\label{fig_err}
\end{figure}

Self-supervised monocular depth estimation relies on the assumption of static scenes and Lambertian surfaces to estimate both depth and relative pose. However, the assumption may not hold in many scenarios, leading to unstable unsupervised learning and local minimum issues in dynamic regions and non-Lambertian or low-textured surfaces. To mitigate this challenge, recent works \cite{hr_depth,ren2022adaptive,poggi2020uncertainty} have introduced distillation techniques to facilitate training. These methods first need to build a well-behaved and sophisticated teacher model, and then the teacher model with frozen weights is used to distill the student model. The entire distillation process requires two stages of training to be completed separately, thus resulting in inefficient training. In addition, multiscale prediction structures are commonly used in dense prediction tasks now. If high-quality multiscale disparities can be generated, which help improve the depth estimation performance. Multiscale disparities are usually obtained by upsampling low-resolution disparity, but upsampling operations lose valuable information and cause disparity misalignment, bringing negative effects \cite{fapn,li2020semantic,han2022brnet}.

\begin{figure*}[!t]
\centering
\includegraphics[width=1.0\textwidth]{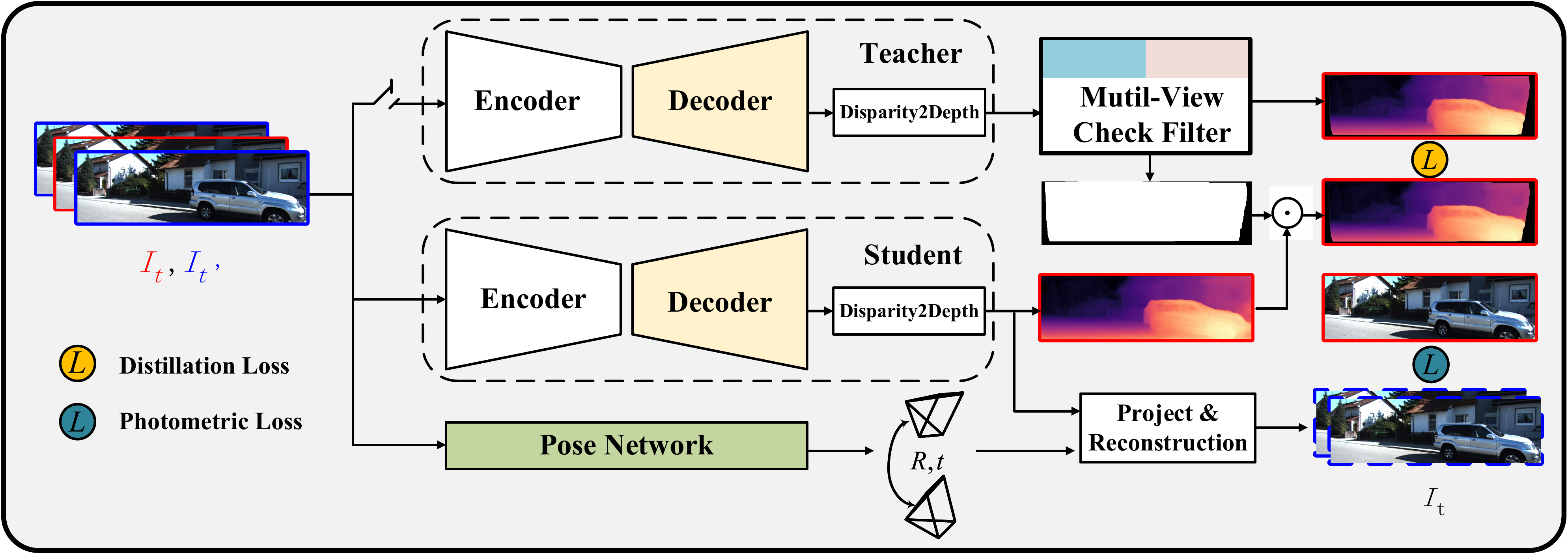}
\caption{Overview of the proposed framework. The model adopts an encoder-decoder architecture, in which the encoder is a commonly used backbone, such as ResNet \cite{resnet} and Swin Transformer \cite{swin_transformer}. Our decoder outputs disparity, which is then converted to depth. The branch of the teacher model provides depth supervision to the student model after outliers are filtered out by a multiview check filter. In the architecture, the input sequence consists of $I_t$ and $I_{t'}$, where $I_{t'}\in\{ I_{t-1},I_{t+1}\}$. $I_t$ is used to output depth and $I_{t'}$ is used for reconstruction to generate $\widetilde{I_{t}}$.}
\label{fig_framework}
\end{figure*}

To address these issues, we propose two novel ideas to enhance self-supervised monocular depth estimation: 1) self-reference distillation and 2) disparity offset refinement. Specifically, to provide additional supervision during the training phase, we train a teacher model with a self-supervised approach during the initial epoch. The teacher model and its distilled counterpart, i.e., the student model, share an identical structure. With the increase of training epochs, the teacher model is continuously updated by inheriting the optimized parameter, so as to provide better depth supervision. However, the depth generated by the teacher model on all pixels is not necessarily reliable, and the edge area and the motion area will have relatively low confidence (Fig.\ref{fig_err}). We do not expect the teacher model to distill this kind of knowledge with large depth errors to the student model. To obtain better supervision signals, we introduce a multiview check filter to filter outliers in the depth map through two steps of forward projection and backprojection of the camera target view and the source view. For the misalignment of multiscale disparity maps, we introduce the disparity offset fields to refine the disparity output by leveraging the contextual consistency between high-level and low-level features. The disparity offset fields allow the multiscale disparities to be aligned and enhance the depth prediction. The results can be found in the ablation experiment in Table \ref{table_ablation}. 

To summarize, the main contributions are listed as follows:
\begin{itemize}
\item We propose a novel self-supervised monocular depth estimation method employing distillation technique and disparity offset refinement to effectively improve the depth estimation performance.
\item We propose self-reference distillation and introduce the multiview check technique to remove the depth outliers from the teacher model, implementing efficient single-stage online distillation learning. 
\item We leverage the contextual consistency in adjacent features to predict the disparity offset field. The aligned refinement for the disparity solves the disparity misalignment problem caused by the upsampling process.
\item We conduct extensive experiments on the KITTI and Make3D datasets, demonstrating that our model outperforms existing state-of-the-art methods.
\end{itemize}

\section{Related Work}
This section reviews the literature on monocular depth estimation and knowledge distillation.
\subsection{Supervised monocular depth estimation}
Depth estimation from a single image is an inherently ill-posed problem as pixels in the image can have numerous plausible depths. Saxena et al. \cite{saxena2005learning} used a discriminatively trained Markov Random Field that incorporates multiscale local and global features and modeled both depths at individual points as well as the relation between depths at different points. Eigen et al. \cite{eigen} introduced a multiscale architecture to make a coarse global depth prediction and progressively refine this prediction locally using two separate networks. A representative BTS method was proposed by \cite{lee2019big} using local planar guidance layers to guide the features to full resolution instead of standard upsampling layers during the decoding phase. Shariq et al. \cite{adabins} divided the depth range into bins whose center value is estimated adaptively per image. The final depth values are estimated as linear combinations of the bin centers. Based on the different features generated by the encoder, Shao et al. \cite{shao2022towards} used the strategy of ensemble learning to obtain more robust depth prediction. Song et al. \cite{song2021monocular} adopted the Laplacian pyramid for resolving the problem of monocular depth estimation. By recovering depth residuals from encoded features in different levels of the Laplacian pyramid and summing up those predicted results progressively. 

Various fully supervised methods based on deep learning have been continuously explored. However, all the above methods require high-quality ground-truth depth, which is costly to obtain.

\begin{figure*}[!t]
\centering
\includegraphics[width=1.0\textwidth]{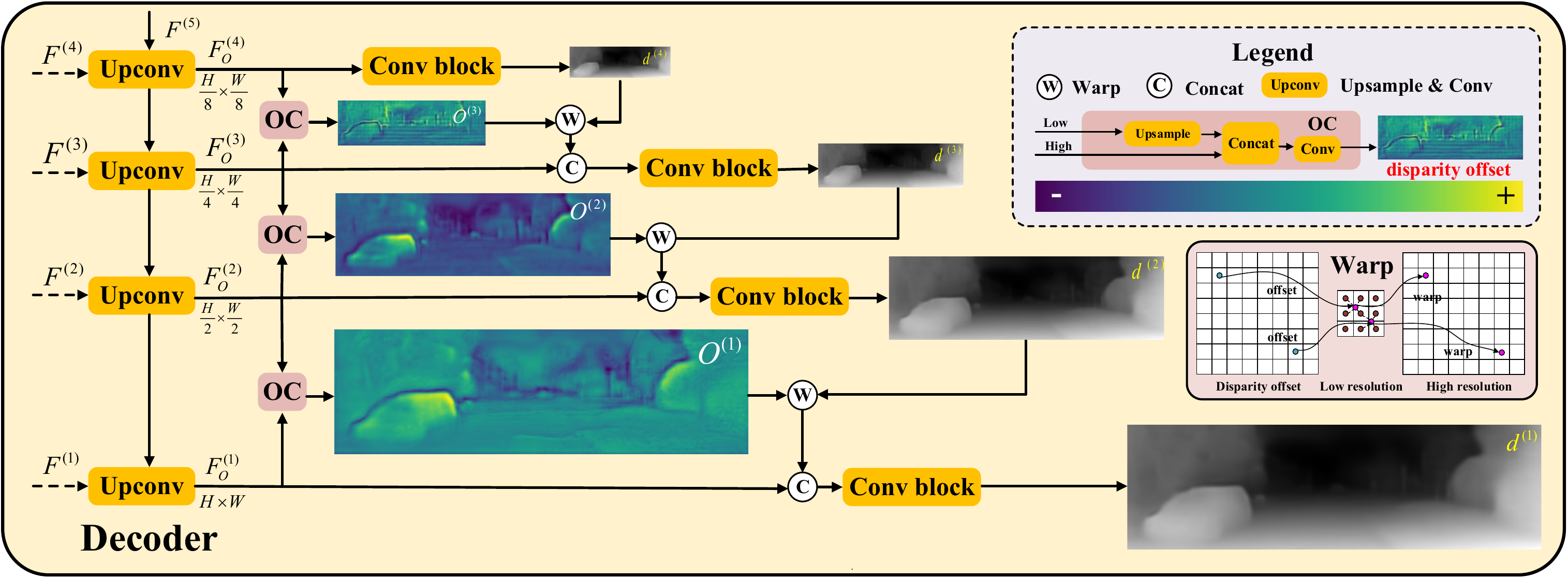}
\caption{Details of the decoder. The input of the decoder comes from the multiscale feature $F^{(i)}$ generated by the encoder, and the output is a disparity map of H$\times$W size. The disparity output by the decoder is obtained by progressively upsampling the low-resolution disparity map. In this process, the offset calculation (OC) module provides a disparity offset field to fuse with the upsampled disparity to achieve disparity alignment. In the warp process, the high-resolution disparity map is the bilinear interpolation of the neighboring pixels in low-resolution disparity map, where the neighborhoods are defined according learned disparity offset.}
\label{fig_decoder}
\end{figure*}

\subsection{Self-Supervised monocular depth estimation}
While fully supervised approaches for depth estimation advance rapidly, the availability of precise depth labels becomes a significant problem. Hence, more recent self-supervised works provide alternatives to avoid the need for ground-truth depth annotations.

In monocular depth estimation, self-supervised approaches unify depth estimation and ego-motion estimation into one framework using view synthesis as a supervision signal. Zhou et al. \cite{zhou} proposed an unsupervised learning framework for monocular depth estimation and camera motion estimation from unlabeled video sequences. Zou et al. \cite{zou2018df} proposed leveraging geometric consistency as additional supervision signals for simultaneously training single-view depth prediction and optical flow estimation models using unlabeled video sequences based on brightness constancy and spatial smoothness priors. Furthermore, Godard et al. \cite{mono2} proposed a classical method Monodepth2, and they adopted an auto-masking scheme to filter out invalid pixels from moving objects and introduced a minimum reprojection loss to address occlusions. Based on Monodepth2, numerous current self-supervised monocular depth estimation approaches \cite{cad,monovit,diffnet} are further researched. Liu et al. \cite{liu2021self} proposed a domain-separated network for self-supervised depth estimation of all-day images. Michael et al. \cite{9578595} presented a novel method for predicting accurate depths by exploiting wavelet decomposition. Shu et al. \cite{chen2021fixing} exploited the point cloud consistency constraint to optimize view synthesis process. Jaehoon et al. \cite{choi2020safenet} exploited semantic-aware depth features that integrate the semantic and geometric knowledge to overcome the limitations of the photometric loss. Vitor et al. \cite{guizilini20203d} leveraged novel symmetrical packing and unpacking blocks to jointly learn to compress and decompress detail-preserving representations using 3D convolutions and implement a self-supervised monocular depth estimation method combining geometry with a new deep network. Akhil et al. \cite{gurram2021monocular} performed monocular depth estimation by virtual-world supervision and real-world SfM self-supervision. They compensate the SfM self-supervision limitations by leveraging virtual-world images with accurate semantic and depth supervision, and addressing the virtual-to-real domain gap. Other published methods were based on feature representation learning \cite{spencer2020defeat}, competitive collaboration \cite{ranjan2019competitive}, edge, normal \cite{yang2018lego,yang2017unsupervised}, semantic segmentation \cite{guizilini2020semantically,klingner2020self}.

\begin{figure*}[!t]
\centering
\includegraphics[width=1.0\textwidth]{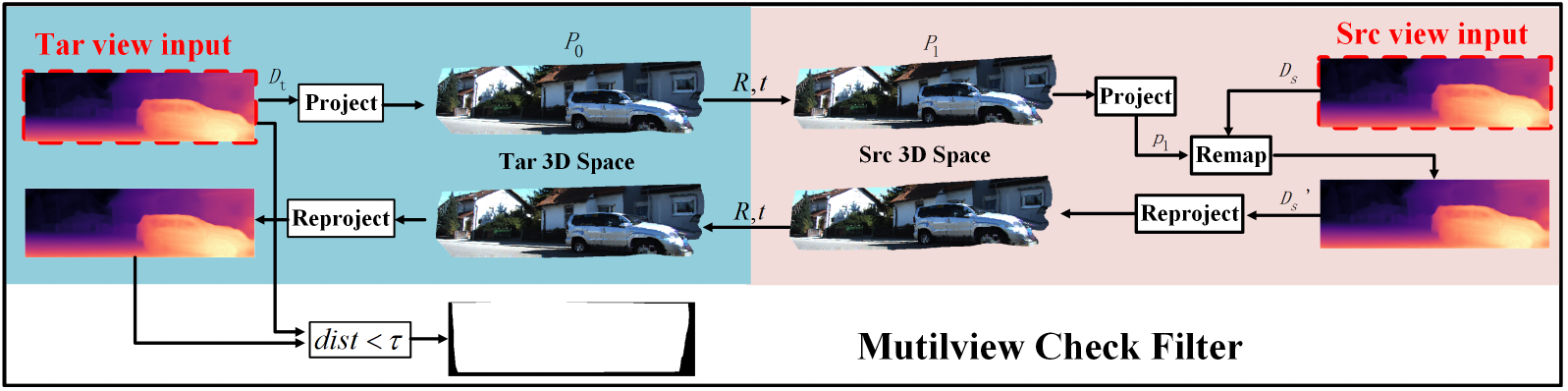}
\caption{Mutilview check filter. This process uses the target view and the source view to perform forward projection and backward reprojection. In the figure, dist represents a certain distance function and pixels exceeding the threshold are filtered out. The white part of the output mask is the reserved part, and the black part is filtered out. The tar view input corresponds to the $Z_{ct}$ in Eq.\ref{eq7}, and the detailed formulation of $dist < \tau$ can be seen in Eq.\ref{eq11}.}
\label{fig_mutilcheck}
\end{figure*}

\subsection{Knowledge Distillation}
The concept of knowledge distillation was first proposed by \cite{distill} and made popular by \cite{hinton2015distilling}. Knowledge distillation aims to transfer knowledge from a teacher model to obtain a powerful and lightweight student model. The idea has been exploited for many computer vision tasks \cite{li2017learning,chen2017learning,gupta2016cross}. 

Recently, some works have attempted to exploit distillation for unsupervised depth estimation. Ren et al. \cite{ren2022adaptive} proposed an adaptive co-teaching framework for unsupervised depth estimation that enjoys the strengths of knowledge distillation and ensemble learning for more accurate depth estimation. Matteo et al. \cite{poggi2020uncertainty} proposed a new and peculiar self-teaching paradigm to model uncertainty and then used the uncertainty to assist knowledge distillation.  Lyu et al. \cite{hr_depth} adopted a large model to improve the accuracy of a lightweight model through two-stage training. These methods usually need to build a complex teacher model. The training of the teacher model and distillation processes are completely separated, thus resulting in relatively large time and computational costs.

Inspired by BAN \cite{furlanello2018born} training a student model similarly parameterized as the teacher model and making the trained student be a teacher model in a new round, we propose our self-reference learning mode in monocular depth estimation. The self-training scheme \cite{xie2020self} generates distillation labels for unlabeled data and trains the student model with these labels. Different from the two-stage distillation method used in previous work \cite{ren2022adaptive,poggi2020uncertainty,hr_depth}, our self-reference distillation achieves efficient single-stage online distillation.

\section{Method}
In this section, we first present the proposed network architecture. Then, we describe the implementation details of self-reference distillation, including the construction of a teacher-student distillation model and the implementation of a multiview check filter. Finally, we provide a detailed account of how we leverage contextual consistency between high-level and low-level features to obtain the disparity offset. 

\subsection{Motivation}
In self-supervised monocular depth estimation, the depth and relative pose are estimated together, and these two intermediate variables are used to perform projection and reprojection operations to synthesize images. Then the photometric error is minimized to train the model. Static scenes and Lambertian surfaces are important underlying assumptions for self-supervised depth estimation. However, dynamic regions, non-Lambertian surfaces or low-texture surfaces violate this assumption, causing unstable unsupervised training and local minima problems. To mitigate this challenge, recent works have introduced distillation techniques to facilitate training. These methods first need to build a well-behaved and sophisticated teacher model, and then the teacher model with frozen weights is used to distill the student model. The entire distillation process requires two stages of training to be completed separately, thus resulting in inefficient training. Therefore, we design efficient single-stage online distillation learning to further alleviate this problem. We first let the model optimize a set of parameters through self-supervision in the first training epoch. In the next epoch of training, the optimized parameters will be loaded for direct inference and generate depth pseudolabels. To produce higher-quality depth supervision, we propose a multiview check filter, which subjects depth pseudolabels to multiple views to filter out outliers.

In addition, multiscale prediction structures are commonly used in dense prediction tasks now. If high-quality multiscale disparities can be generated, it will help improve the performance of depth estimation. Multiscale disparities are usually obtained by upsampling low-resolution disparity, but upsampling operations will lose valuable information and cause disparity misalignment, bringing negative effects \cite{fapn,li2020semantic,han2022brnet}. Therefore, we leverage the contextual consistency between high-level and low-level features to obtain the disparity offset and refine the disparity output incrementally based on the disparity offset to align disparity information at different scales.

\subsection{Network Architecture}
For self-supervised monocular depth estimation, the proposed method utilizes an encoder-decoder architecture with skip connections. In this architecture, the encoder transforms the input image into a latent space representation, while the decoder reconstructs the disparity from the latent space representation, as shown in Fig.\ref{fig_framework}. After converting from disparity to depth, the output of the decoder becomes the model's final output. During the training process, the depth output by the depth network and the relative pose output by the pose 
\begin{figure}[!h]
\centering
\includegraphics[width=0.48\textwidth]{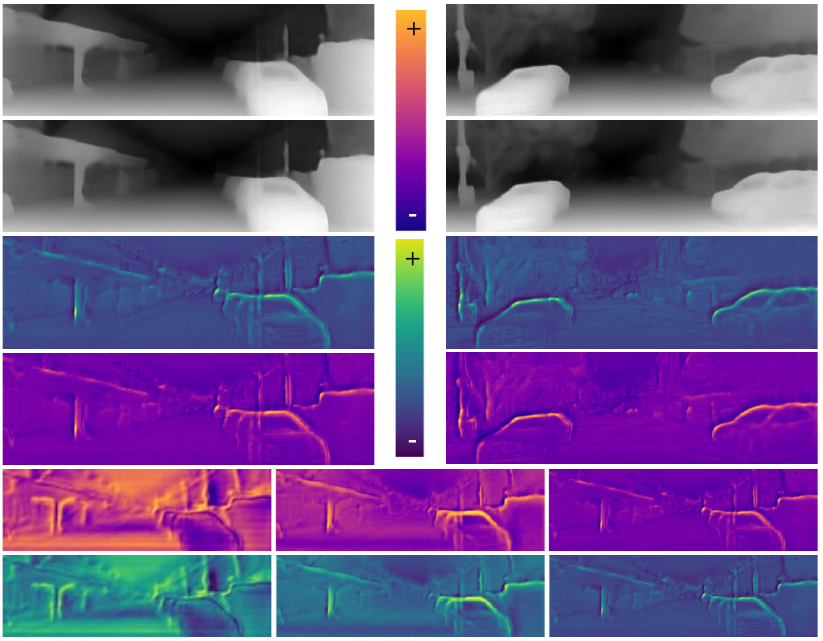}
\caption{Visualization of disparity offset. The first two rows are low-resolution disparity and high-resolution disparity. The third row and the fourth row are horizontal disparity offset and vertical disparity offset. The last two rows are different levels of disparity offset, and from left to right are high level to low level.}
\label{fig_offset}
\end{figure}
network are used for view synthesis, and self-supervised training is achieved by optimizing the photometric loss. Additionally, the output depth of the teacher depth network is used as a pseudolabel to distill knowledge to the student model.

\textbf{Encoder}. The encoder plays a crucial role in effectively extracting features. Therefore, inspired by the outstanding performance of Transformers\cite{swin-depth,yang2021transformer,ranftl2021vision,wang2021pyramid} in various vision tasks, we adopt an improved multipath Vision Transformer architecture MPViT \cite{mpvit}, which leverages both the local connectivity of convolutions and the global context of the transformer. The encoder receives a single frame with resolution $H\times W$ as input and extracts features at five different scales, with resolutions $H\times W$, $\frac{H}{2}\times \frac{W}{2}$,  $\frac{H}{4}\times \frac{W}{4}$, $\frac{H}{8}\times \frac{W}{8}$, and $\frac{H}{16}\times \frac{W}{16}$. These multiscale features are then directly accessed by the decoder through skip connections. According to the differences in layers and channels in the four scales, MPViT \cite{mpvit} has tiny, xsmall, small and base versions. We use MPViT-S (small) with a parameter size of 22.8M, equivalent to ResNet50 (25M)and Swin Transformer-T (28M). The layers of the 4 scales are set to [1,3,6,3], and the channels are set to [64, 128, 216, 288]. It should be noted that we actually use features of 5 scales, so we modify MPViT and add $\frac{H}{2}\times \frac{W}{2}$ scale features. The layers of this scale are set to 3, and the channels are set to 64.

\textbf{Decoder}. The multiscale features output by the encoder are used as input to the decoder, which progressively upsamples and convolves the feature maps to increase their resolution. The upsampled feature maps are then fused with the skip-connected encoding features and passed through the layers in a top-down manner, combining strong semantic information features and high-resolution features. Based on the features of two adjacent scales, the offset calculation module calculates the disparity offset field at each scale. The low-resolution disparity is refined with the assistance of the disparity offset and is progressively passed to the high-resolution scale to obtain the output disparity of the decoder. More specific details on how the offset calculation (OC) module calculates the disparity offset field from the features of adjacent scales will be discussed in the section on disparity alignment. The structure of the depth decoder is shown in Fig.\ref{fig_decoder}. The multi-scale features generate three scale disparity offsets ($O$) in the figure. The final disparity output of the model is generated from the low-resolution disparity step-by-step upsampling, and the disparity map after each upsampling operation will be aligned according to $O$. After the disparity of the H$\times$W scale is refined, the decoder outputs the disparity and then converts it to depth.

\textbf{Pose Network}. Our pose network uses the same lightweight architecture ResNet18\cite{resnet} as \cite{mono2,hr_depth,yan2021channel,zhou2021self}, taking a sequence of three frames as input to predict a 6-DoF relative pose between adjacent frames.

\subsection{Self-Supervised Learning}
\textbf{Disparity alignment}. Dense prediction tasks achieve a better performance with multiscale predictions. In depth estimation, multiscale disparities are usually obtained by upsampling low-resolution disparity. However, commonly used upsampling operations (e.g., bilinear interpolation) lose valuable information and cause disparity misalignment \cite{fapn,li2020semantic,han2022brnet}.

Inspired by various feature alignment works \cite{fapn,alignseg,shallow}, we design an offset calculation module (OC) that utilizes the adjacent scale features to calculate the disparity offset field. The OC outputs allow effective disparity alignment of different scales and improve the prediction accuracy, which is provided evidence in the ablation study.

The OC module in the decoder receives two resolution features\{$F^{(i)}_o,F^{(i+1)}_o$\}. (1) The low-resolution feature $F^{(i+1)}_o$ is first upsampled to obtain features of the same resolution as $F^{(i)}_o$. (2) The resulting features are concatenated with the high-resolution feature $F^{(i)}_o$ to produce the disparity offset $O^{(i)}$, as shown in the offset calculation structure of the legend in Fig.\ref{fig_decoder}. (3) The disparity $d^{(i+1)}$ is warped by $O^{(i)}$ to obtain the refined disparity map (the high-resolution disparity map). In the warp process, the high-resolution disparity map is the bilinear interpolation of the neighboring pixels in low-resolution disparity map, where the neighborhoods are defined according learned disparity offset as shown in Fig.\ref{fig_decoder}. (4) Refined disparity is concatenated with $F^{(i)}_o$ to generate $d^{(i)}$. This process can be mathematically formulated as follows:
\begin{align}
\label{eq1}
O^{(i)} &= Conv(Cat(F_{o}^{(i)},Upsample(F_{o}^{(i+1)}))) ,  \\
d^{(i)} &= Conv(Cat(F^{(i)}_o,Warp(O^{(i)},d^{(i+1)}))),
\end{align}
In the legend of the figure, we mark that the disparity offset is positive or negative, which represents the direction of the disparity offset. The disparity offset includes the horizontal offset and vertical offset. In Fig.\ref{fig_decoder}, we only indicate the horizontal offset for convenience. 

\begin{figure*}[h]
\centering
\includegraphics[width=0.98\textwidth]{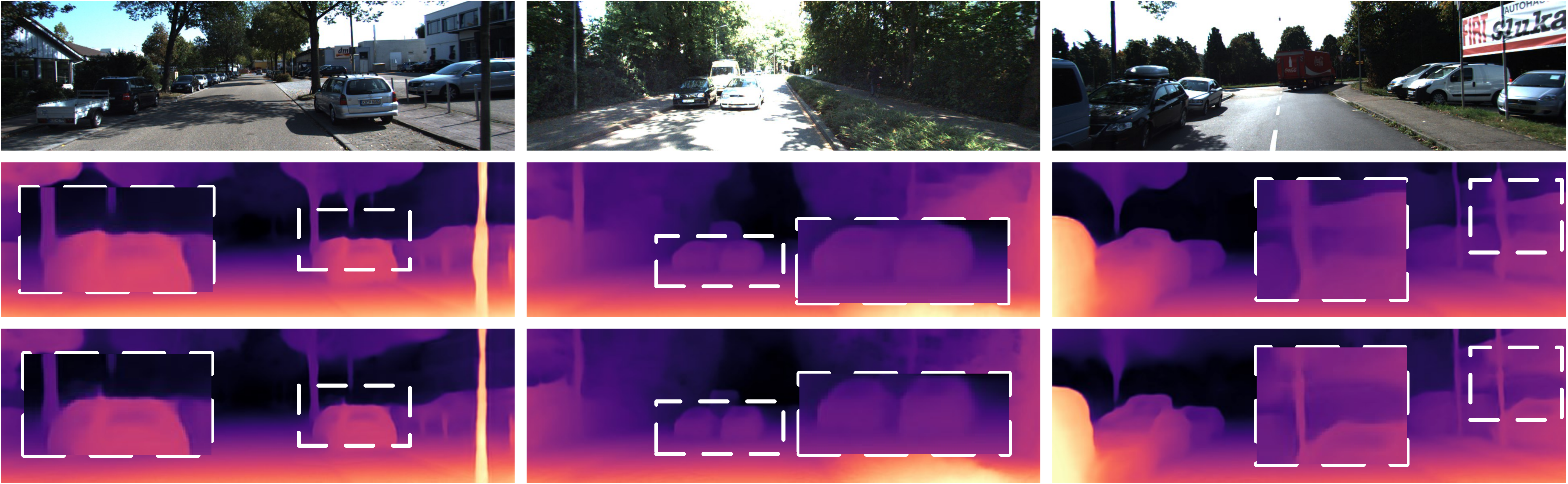}
\caption{Visualization results of w/ DA and w/o DA. From the top to bottom, the RGB images, the predicted depth maps without DA (disparity alignment) and the predicted depth maps with DA. The area inside the white box clearly shows that the depth prediction performance is better with DA than without DA.}
\label{fig_da}
\end{figure*}

In the visualization of Fig.\ref{fig_offset}, we show the disparity offset in the horizontal and vertical directions. The first two rows are low-resolution disparity and high-resolution disparity, respectively and the low-resolution disparity map needs to be aligned with the high-resolution disparity map by offsetting both horizontally and vertically. The third row and the fourth row are the horizontal disparity offset and vertical disparity offset. The last two rows are different levels of disparity offset, and from left to right are high level to low level. As the upsampling proceeds, high level features are propagated to appropriate high-resolution positions following the guidance of disparity offset. Disparity offset have coarse-to-fine trends from high level to low level, so at the low level, the main salient regions are mainly located at the edges with rich details. In the ablation experiments, we verify the effectiveness of the disparity alignment and show the visualization results of predicted depth with disparity alignment and without disparity alignment in Fig.\ref{fig_da}. Our proposed DA is much clearer in the estimation of edge contours, which is consistent with the performance of the multiscale disparity offsets.

\textbf{Photometric Loss}. Self-supervised monocular depth estimation performs image synthesis by minimizing the photometric loss during training, so that the images synthesized from other views are close to the appearance of the target image, and the depth and relative pose required for the synthesized image are solved during this optimization process. $I_{t}$ denotes the target image and $I_{t'}$ denotes the source image i.e. $I_{t'}\in\{ I_{t-1},I_{t+1}\}$. The relative pose of each source image's associated image is indicated as $T_{t\rightarrow{t'}}$ and the camera intrinsics are denoted as $K$. $D_t$ is a depth map transformed from the disparity $d^{(1)}$. Similar to \cite{mono2,mono,garg}, we predict a dense depth map that minimizes the photometric reprojection loss $\mathcal{L}_{pe}$, where
\begin{align}
\label{eq3}
\widetilde{I_{t}} &= \pi(I_{t'},K,D_t,T_{t\rightarrow{t'}}) ,        \\
\mathcal{L}_{pe} &= min \mathcal{F}(I_{t},\widetilde{I_{t}}),  \\
\mathcal{F}(I_{t},\widetilde{I_{t}}) &= \alpha \dfrac{1-SSIM(I_{t},\widetilde{I_{t}})}{2} + 2\alpha |I_t-\widetilde{I_{t}}|,
\end{align}
Here $\pi$ is a reconstruction function following \cite{mono2,mono} and $\widetilde{I_{t}} \in \{ I_{t-1\rightarrow{t}},I_{t+1\rightarrow{t}}\}$. $\mathcal{F}$ is the weighted sum of the intensity difference term $\mathcal{L}_1$ \cite{mono2,mono} and the structural similarity term $SSIM$ \cite{ssim} and  $\alpha$ is set to 0.85 as \cite{mono2}.

\textbf{Smoothness Loss}. As in \cite{mono2,mono}, we use edge-aware smoothness loss to encourage the smoothness property of inverse depth map:
\begin{align}
\label{eq6}
\mathcal{L}_s &= |\partial_xd^*_t|e^{-|\partial_xI_t|} + |\partial_yd^*_t|e^{-|\partial_yI_t|},        
\end{align}
where $d^*_t = d_t/\Bar{d_t}$ is the mean-normalized inverse depth from \cite{wang2018learning} to discourage shrinking of the estimated depth.

\begin{figure}[!t]
\centering
\includegraphics[width=0.46\textwidth]{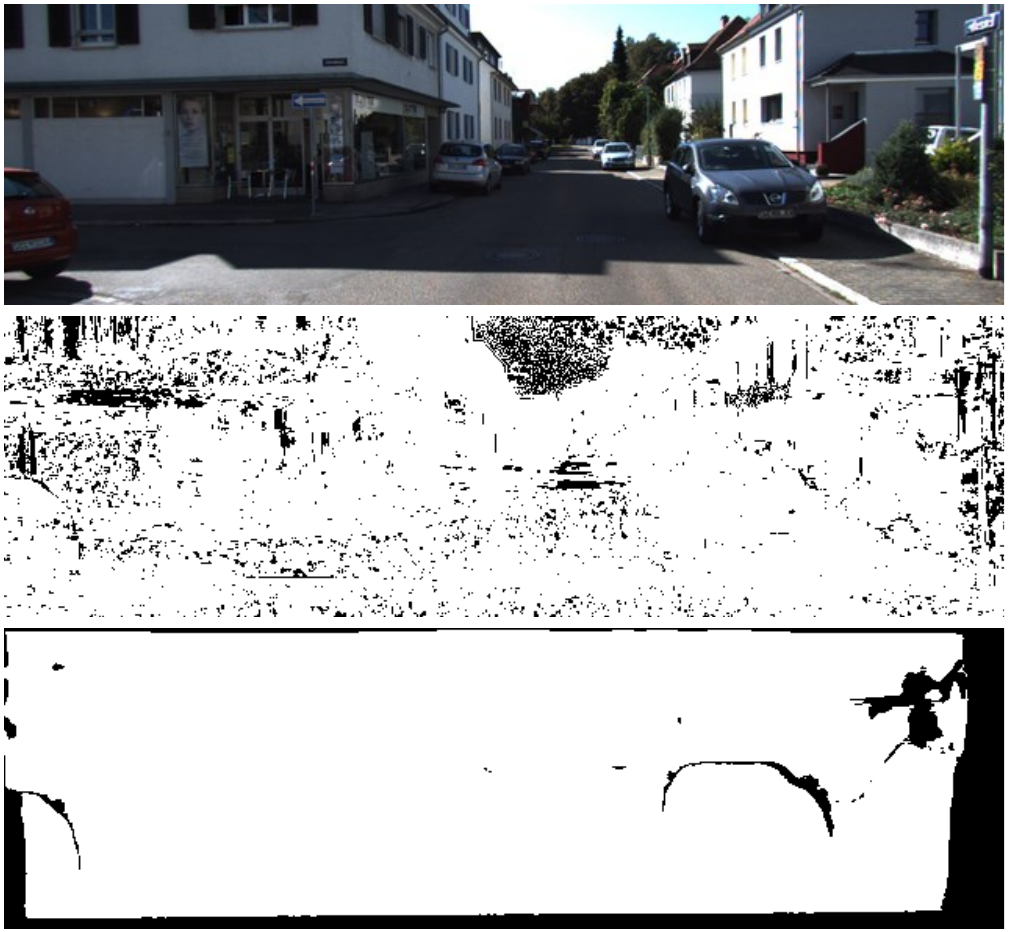}
\caption{Mask comparison. The second row is the auto mask, where the area with large photometric errors is masked and the third row is our filter mask, where the area with large depth errors is masked.}
\label{fig_mask_compare}
\end{figure}

\subsection{Knowledge Distillation}
\textbf{Self-reference distillation}. To mitigate the unstable unsupervised training and local minima problems, knowledge distillation is introduced in depth estimation \cite{hr_depth,ren2022adaptive,poggi2020uncertainty}, using the pseudo depth labels generated by a teacher model as supervision for a student model. 

Our knowledge distillation process is finished in a single stage, online, as opposed to earlier work \cite{petrovai2022exploiting}, which needs a two-stage training procedure: (1) finishing training a well-behaved and sophisticated teacher model and (2) the teacher model with frozen weights is used to distill the student model. In our method, the student branch of Fig.\ref{fig_framework} depicts the model going through self-supervised learning first. The model's weights are saved after the initial training epoch. The weights from the previous epoch of training are loaded online and utilized for inference starting with the second training epoch, which corresponds to the teacher branch in Fig.\ref{fig_framework}. The student model is then distilled using the output of this inference as pseudo depth labels. In terms of network architecture, the teacher branch model is identical to the student branch model, with the only difference being the model weights. As shown in Fig.\ref{fig_framework}, the decoder performs a disparity-to-depth conversion after outputting the disparity. To make more comprehensive use of information, we not only convert the depth of the H$\times$W scale, but also convert the multiscale disparities into multiscale depth maps, which can provide more supervision information to distill the student model on multiple scales. Our distillation loss can be written as: 
\begin{equation}
\label{eq_L_depth}
\mathcal{L}_{d} = \dfrac{1}{4}\sum_{i=1}^4(||D_{teacher}-\hat{D}_{student}||_1)_i.
\end{equation}

\textbf{Mutilview Check}. Although teacher-student distillation provides supervision during training, the prediction of the teacher model at each pixel is not highly reliable. For example, the edge area will have relatively low confidence, resulting in some incorrect outlier points.

Like Monodepth2, we also employ masking strategy to boost performance. The auto-masking strategy of Monodepth2 calculates the minimum photometric error between RGB images to obtain the mask, which is used to re-weight the self-supervised loss term. In knowledge distillation, our aim is to remove out these points with large depth errors. The auto-masking technique does enhance performance by reducing motion scenes, however it masks out points with large photometric errors rather than large depth errors. The area with large photometric errors does not completely reflect the large depth errors as illustrated in Fig.\ref{fig_mask_compare}. Our needs cannot be satisfied by the auto masking strategy.

 To achieve better distillation of depth information, we design a multiview check filter demonstrated in Fig.\ref{fig_mutilcheck}, to filter outliers and offer a hard mask for teacher-student distillation. Specifically, the multiview check filter receives the depth output from the model, which consists of the depth map from the target view and the depth map from the source view. According to the imaging principles of a pinhole camera, the correspondence between the pixel points of each perspective and the 3D spatial points can be mathematically described:

\begin{align}
\label{eq_pinhole}
Z_{c}\begin{bmatrix} u \\ v  \\ 1 \end{bmatrix}
&=  \begin{bmatrix}\mathbf{K}&\mathbf{0}\end{bmatrix}
\begin{bmatrix}\mathbf{R} &\mathbf{t}\\ 
0 &1 \end{bmatrix}
\begin{bmatrix} X_{w} \\ Y_{w}  \\ Z_{w}  \\1 \end{bmatrix},\\
&= \begin{bmatrix}\mathbf{K}&\mathbf{0}\end{bmatrix}
\begin{bmatrix} X_{c} \\ Y_{c}  \\ Z_{c} \\1 \end{bmatrix},\\
\mathbf{K} &= \begin{bmatrix} f_x &0& u_0 \\ 0& f_y &v_0  \\ 0 &0 &1 \end{bmatrix},
\end{align}
where $\mathbf{K}$ refers to camera intrinsic matrix, $\begin{bmatrix}\mathbf{R} &\mathbf{t}\\ 
0 &1 \end{bmatrix}$ refers to camera extrinsic matrix. $(X_w,Y_w,Z_w)$ and $(X_c,Y_c,Z_c)$ represent points in the world coordinate and camera coordinate, respectively.

We use $\mathbf{p_0}=(u,v)$ to represent an arbitrary point in the target view and project it into the 3D space of the target view 
 $\mathbf{P_0}=(X_{ct},Y_{ct},Z_{ct})$ through depth $\mathbf{D_t}(u,v) = Z_{ct}$, as shown in Eq.\ref{eq7}.
\begin{align}
\label{eq7}
\begin{bmatrix} X_{ct} \\ Y_{ct}  \\ Z_{ct}  \end{bmatrix}
&=  \mathbf{K}^{-1}Z_{ct}\begin{bmatrix} u \\ v  \\ 1 \end{bmatrix},
\end{align}
Combined with the relative pose $T_{t\rightarrow{t'}}$ output by the pose network, we can obtain the 3D representation $\mathbf{P_1}=(X_{cs},Y_{cs},Z_{cs})$ of $\mathbf{P_0}$ in the source view, as shown in Eq.\ref{eq8_1}.
\begin{equation}
\label{eq8_1}
\begin{bmatrix} X_{cs} \\ Y_{cs}  \\ Z_{cs}  \end{bmatrix} = 
T_{t\rightarrow{t'}} \begin{bmatrix} X_{ct} \\ Y_{ct}  \\ Z_{ct}  \end{bmatrix}
= T_{t\rightarrow{t'}}\mathbf{K}^{-1} Z_{ct}\begin{bmatrix} u \\ v  \\ 1 \end{bmatrix},
\end{equation}
The $\mathbf{P_1}$ is projected to the 2D point $\mathbf{p_1}=(u_1,v_1)$ of the source view through $Z_{cs} = {\mathbf{D_s}} = model(I_{t'})$ as shown in Eq.\ref{eq7_1}.
\begin{align}
\label{eq7_1}
Z_{cs}\begin{bmatrix} u_1 \\ v_1  \\ 1 \end{bmatrix}
&=  \mathbf{K}\begin{bmatrix} X_{cs} \\ Y_{cs}  \\ Z_{cs}  \end{bmatrix},
\end{align}
Then, the source view depth map $\mathbf{D_s}$ is remapped to $\mathbf{D_s'}$ according to $\mathbf{p_1}=(u_1,v_1)$. Similar to the projection forward process mentioned above, we use $\mathbf{D_s'}$ to perform a reprojection process to obtain reprojected 3D point $\mathbf{\widetilde{P_0}}$, 2D point $\mathbf{\widetilde{p_0}}$ and depth map $\mathbf{\widetilde{D_r}}$ in the target view.

\begin{figure}[h]
\centering
\includegraphics[width=0.48\textwidth]{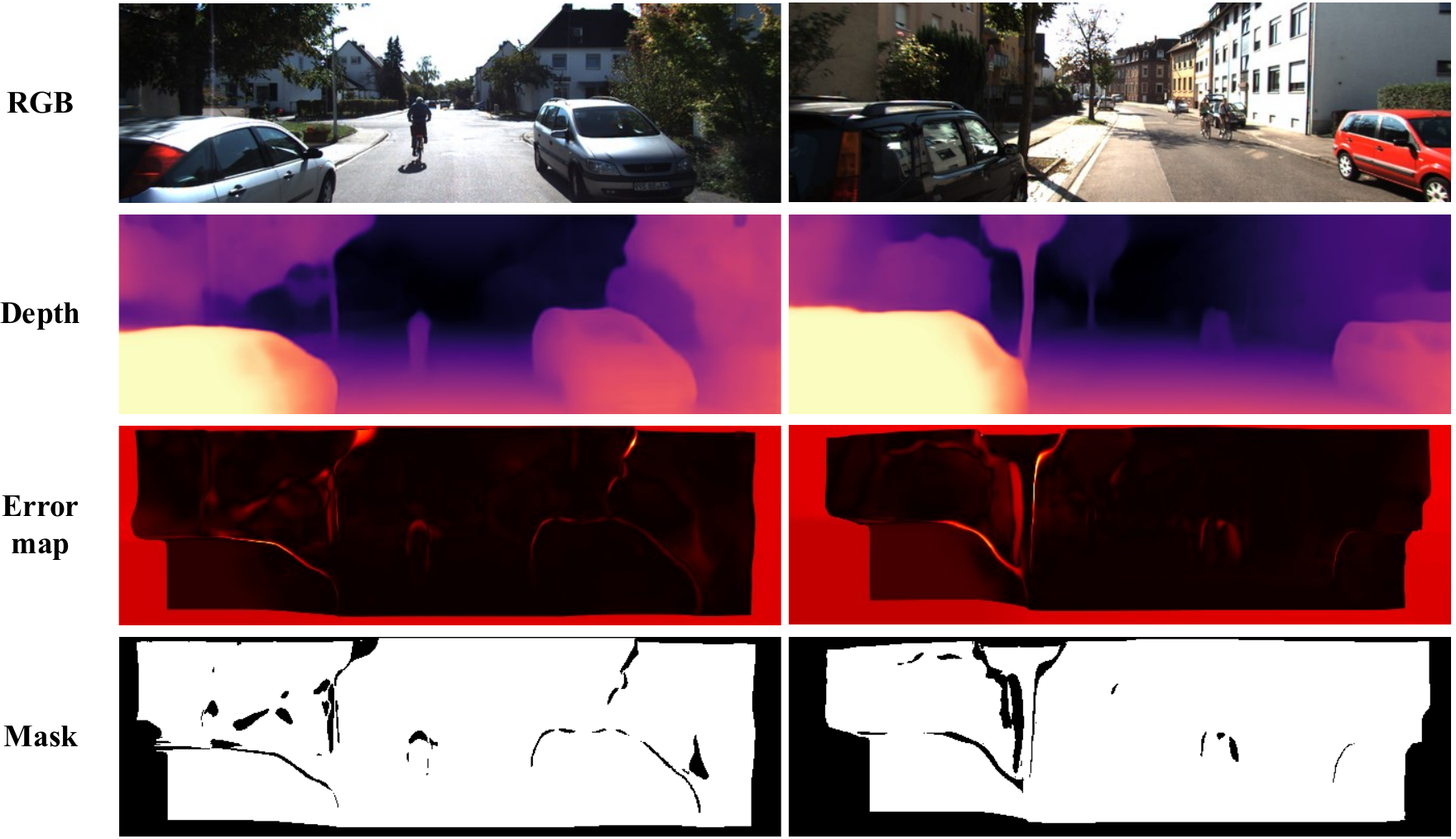}
\caption{Visualization of the filter mask. In the fourth row, the black area is filtered out, and the white area is retained.}
\label{fig_err2}
\end{figure}

A reprojection error at $\mathbf{\widetilde{p_0}}$ is defined as  $e_{reproj}$, and a geometric error $e_{geo}$ is used to measure the relative depth error, which is formulated as \cite{yao2018mvsnet}:
\begin{align}
\label{eq8}
e_{reproj} &= ||\mathbf{\widetilde{p_0}}-\mathbf{p_0}||_2, \\
e_{geo} &= \dfrac{|\mathbf{D_r}(\mathbf{p_0})-\mathbf{D_r}(\mathbf{\widetilde{p_0}})|}{\mathbf{D_r}(\mathbf{p_0})},
\end{align}
The valid subset of pixels for the filter mask is determined by Eq.\ref{eq11}, where $\alpha$ and $\beta$ are hyperparameters. We set $\alpha$ = $\beta$ = 4. 
\begin{align}
\label{eq11}
\{\mathbf{p_0} \}_i= \{\mathbf{p_0} | e_{reproj}<\alpha \Bar{e}_{reproj}, e_{geo}<\beta \Bar{e}_{geo}\},
\end{align}
$\{\mathbf{p_0} \}_i$ represents the set of multiview checks between the target view and the $i$-th source view, and the intersection of the sets calculated under all source views is the filter mask. We show the $e_{geo}$ and the filter mask in Fig.\ref{fig_err} and Fig.\ref{fig_err2}. In Fig.\ref{fig_err2}, the error map on the third row is $e_{geo}$, and the mask on the fourth row is the filter mask. In the error map, darker areas represent smaller relative errors (black areas), and lighter areas have greater errors (red, yellowish areas). In knowledge distillation, it is not expected that teacher model to distill inaccurate depth information to the student model. Inaccurate depth information is mainly distributed on the border around the image and a small part is distributed in the middle area as shown in Fig.\ref{fig_err2}. These outliers with relatively large errors are harmful to the student model if they also participate in knowledge distillation. The most direct filtering method is to use Eq.\ref{eq11} to generate a binary hard mask. It is easy to find that the filter mask and error map in Fig.\ref{fig_err2} may not necessarily correspond exactly. In the picture on the left, the points in the yellow area (with relatively large errors) are basically filtered out, while in the picture on the right, only part of the points in the yellow area are filtered out. This is caused by two reasons: one is caused by the binary hard mask, which will be filtered out only when it reaches a certain threshold, and the other is related to the setting of the threshold. Different image scenes require different thresholds ($\alpha$ and $\beta$) for truncation, but for ease of implementation, we choose a unified $\alpha$ and $\beta$.

In fact, we also consider using a soft mask as a weight to balance the supervision loss term. We use the error map directly as a soft mask, i.e. $M = 1- e_{geo}$ to aid the distillation. However, the soft masking method impairs the distillation. Because the supervisory signal is not truncated in the part with large error, this method also brings negative optimization, as shown in Table \ref{table_ablation} of the ablation experiment results.

\textbf{Distillation Loss}. After the depth pseudo labels produced by the teacher model are filtered by a multiview check filter, we apply the resulting filter masks to our distillation loss. The modified distillation loss can be formulated as: 
\begin{align}
\label{eq12}
\mathcal{L}_{d} = \dfrac{1}{4}\sum_{i=1}^4(M||D_{teacher}-\hat{D}_{student}||_1)_i,
\end{align} 
where $M$ is the filter mask.

\textbf{Total Loss}
The total loss in training consists of three parts, photometric loss $\mathcal{L}_{pe}$, smoothing loss $\mathcal{L}_{s}$ and distillation loss$\mathcal{L}_{d}$, which are calculated at four scales.
\begin{align}
\label{eq13}
\mathcal{L}_{total} = \dfrac{1}{4}\sum_{i=1}^4(\mu\mathcal{L}_{pe}+\lambda\mathcal{L}_{s}+\gamma\mathcal{L}_{d})_i,
\end{align} 
with $\lambda$ set to $10^{-3}$ and $\gamma$ set to $0.1$. Similar to previous works\cite{mono2,monovit}, we apply a per-pixel binary mask, i.e. $\mu \in\{0,1\}$, which is formulated as:
\begin{align}
\label{eq14}
\mu = [min \mathcal{F}(I_{t},\widetilde{I_{t}})<min \mathcal{F}(I_{t},I_{t'})],
\end{align} 
where [] is the Iverson bracket.

\begin{table*}[h]
    \renewcommand{\arraystretch}{1.2}
    \centering
    \newcolumntype{L}[1]{>{\raggedright\let\newline\\\arraybackslash\hspace{0pt}}m{#1}}
    \newcolumntype{C}[1]{>{\centering\let\newline\\\arraybackslash\hspace{0pt}}m{#1}}
\newcolumntype{R}[1]{>{\raggedleft\let\newline\\\arraybackslash\hspace{0pt}}m{#1}}
    \resizebox{2.0\columnwidth}{!}{
    \begin{tabular}{L{35mm} | C{8mm}||C{13mm} |C{10.5mm} C{10.5mm} C{10.5mm} C{13mm}| C{13mm} C{13mm} C{13mm}}
        \toprule
        Method & Train & Backbone & Abs Rel & Sq Rel & RMSE & RMSE log & $\delta<1.25$ & $\delta<1.25^2$ & $\delta<1.25^3$\\
        \hline
        Eigen \cite{eigen} & D & - & 0.203 & 1.548 &  6.307 & 0.282 & 0.702 &0.890 & 0.890 \\
        Liu \cite{liu} & D & VGG16 & 0.201 & 1.584 & 6.471 & 0.273 & 0.680 &0.898 & 0.967 \\
        Klodt \cite{klodt2018supervising} & D*M & ResNet50 & 0.166 & 1.490 &  5.998 & - & 0.778 &0.919 & 0.966 \\
        AdaDepth \cite{kundu2018adadepth} & D* & ResNet50 & 0.167 & 1.257 &  5.578 & 0.237 & 0.771 &0.922 & 0.971 \\
        DVSO \cite{yang2018deep} & DS & ResNet50 & 0.097 & 0.734 &  4.442 & 0.187 & 0.888 & 0.958 & 0.980 \\
        SVSM FT \cite{luo2018single} & DS & VGG16 & \textbf{0.094} & \textbf{0.626} &  4.252 & 0.177 & 0.891 & 0.965 & 0.984 \\
        Guo \cite{guo2018learning} & DS & VGG16 & 0.096 & 0.641 &  \textbf{4.095} & \textbf{0.168} & \textbf{0.892} &\textbf{0.967} & \textbf{0.986} \\
         \hline
         \hline 
        Monodepth2 \cite{mono2} & M & ResNet18 & 0.115 & 0.903 &  4.863 & 0.193 & 0.877 &0.959 & 0.981 \\
        MonoDEVSNet \cite{gurram2021monocular} & M & ResNet18 & 0.116 & 0.836 &  4.735 & - & 0.860 &0.954 & - \\
        R-MSFM3 \cite{r_msfm} & M & ResNet18 & 0.114 & 0.815 &  4.712 & 0.193 & 0.876 &0.959 & 0.981 \\
        R-MSFM6 \cite{r_msfm} & M & ResNet18 & 0.112 & 0.806 &  4.704 & 0.191 & 0.878 &0.960 & 0.981 \\
        SAFENet \cite{choi2020safenet} & M+Se & ResNet18 & 0.112 & 0.788 &  4.582 & 0.187 & 0.878 &0.963 & 0.983 \\
        VC-Depth \cite{zhou2020constant} & M & ResNet18 & 0.112 & 0.816 &  4.715 & 0.190 & 0.880 &0.960 & 0.982 \\ 
        \textbf{Ours} & M & ResNet18 & 0.111 & 0.762 & 4.619 & 0.186 & 0.877 & 0.961 & 0.983 \\ 
        \hline
        Shu \cite{chen2021fixing}& M & ResNet50 & 0.129 & 0.976 &  4.958 & 0.203 & 0.848 & 0.951 & 0.979 \\
        Mono-Uncertainty \cite{poggi2020uncertainty}& M & ResNet50 & 0.111 & 0.863 &  4.756 & 0.188 & 0.881 & 0.961 & 0.982 \\
        PackNet\dag \cite{guizilini20203d}& M & PackNet & 0.108 & 0.727 &  4.426 & 0.184 & 0.885 &0.963 & 0.983 \\
        Johnston et al. \cite{johnston2020self} & M & ResNet101 & 0.106 & 0.861 &  4.699 & 0.185 & 0.889 & 0.962 & 0.982 \\
        MonoFormer \cite{bae2022monoformer} & M & Res50+ViT & 0.106 & 0.839 &  4.627 & 0.185 & 0.884 & 0.962 & 0.983 \\
        CADepth \cite{cad} & M & ResNet50 & 0.105 & 0.769 &  4.535 & 0.181 & 0.892 & 0.964 & 0.983 \\
        \textbf{Ours} & M & ResNet50 & 0.106 & 0.718 & 4.520 & 0.180 & 0.886 & 0.964 & 0.983 \\
        \hline
        DIFFNet \cite{diffnet} & M & HRNet18 & 0.102 & 0.749 &  4.445 & 0.179 & 0.897 & 0.965 & 0.983 \\
        MonoViT \cite{monovit} & M & MPViT-S & \textbf{0.099} & 0.708 &  4.372 & 0.175 & \textbf{0.900} &\textbf{0.967} & 0.984 \\
        \textbf{Ours} & M & MPViT-S & \textbf{0.099}& \textbf{0.659}& \textbf{4.314}& \textbf{0.174}& 0.898 & \textbf{0.967}& \textbf{0.985} \\
        \hline
        \hline
        Monodepth2(1024$\times$320) \cite{mono2} & M & ResNet18 & 0.115 & 0.882 &  4.701 & 0.190 & 0.879 &0.961 & 0.982 \\
        R-MSFM3(1024$\times$320) \cite{r_msfm} & M & ResNet18 & 0.112 & 0.773 &  4.581 & 0.189 & 0.879 &0.960 & 0.982 \\
        R-MSFM6(1024$\times$320) \cite{r_msfm} & M & ResNet18 & 0.108 & 0.748 &  4.470 & 0.185 & 0.889 &0.963 & 0.982 \\
        \textbf{Ours(1024$\times$320)} & M & ResNet18 & 0.106 & 0.673 &  4.379 & 0.180 & 0.886 &0.965 & 0.984 \\
        \hline     
        DCNDepth(1024$\times$320) \cite{masoumian2021gcndepth} & M & ResNet50 & 0.104 & 0.720 &  4.494 & 0.181 & 0.888 & 0.965 & 0.984 \\ 
        CADepth(1024$\times$320) \cite{cad} & M & ResNet50 & 0.102 & 0.734 &  4.407 & 0.178 & 0.898 & 0.966 & 0.984 \\
        \textbf{Ours(1024$\times$320)} & M & 
        ResNet50 & 0.102 & 0.653  & 4.381 & 0.178 & 0.898 & 0.966 & \textbf{0.985} \\
        \hline
        DIFFNet(1024$\times$320) \cite{diffnet} & M & HRNet18 & 0.097 & 0.722 &  4.345 & 0.174 & 0.907 & 0.967 & 0.984 \\
        MonoViT(1024$\times$320) \cite{monovit} & M & MPViT-S & \textbf{0.096} & 0.714 &  4.292 & 0.172 & \textbf{0.908} & 0.968 & 0.984 \\
        \textbf{Ours}(1024$\times$320) & M & MPViT-S & \textbf{0.096}& \textbf{0.635}& \textbf{4.158}&  \textbf{0.171}& 0.905&   \textbf{0.969}& \textbf{0.985} \\
        \bottomrule     
    \end{tabular}}
    \vspace{1em}
    \caption{\textbf{Quantitative results}. 
 {\upshape Comparison of our method to existing methods on KITTI 2015 \cite{kitti} using the Eigen split. The best results in each category are in bold. The resolutions we used for training and testing the models were 640$\times$192 and 1024$\times$320 (marked in the table). In the training approach, Se stands for training with semantic labels, D for depth supervision, D* for auxiliary depth supervision, and M for mono self-supervision. $\dag$ refers to the model pretained on Cityscapes \cite{cordts2016cityscapes}.}}
     \label{table_1}
\end{table*}

\begin{figure*}[!t]
\centering
\includegraphics[width=1.0\textwidth]{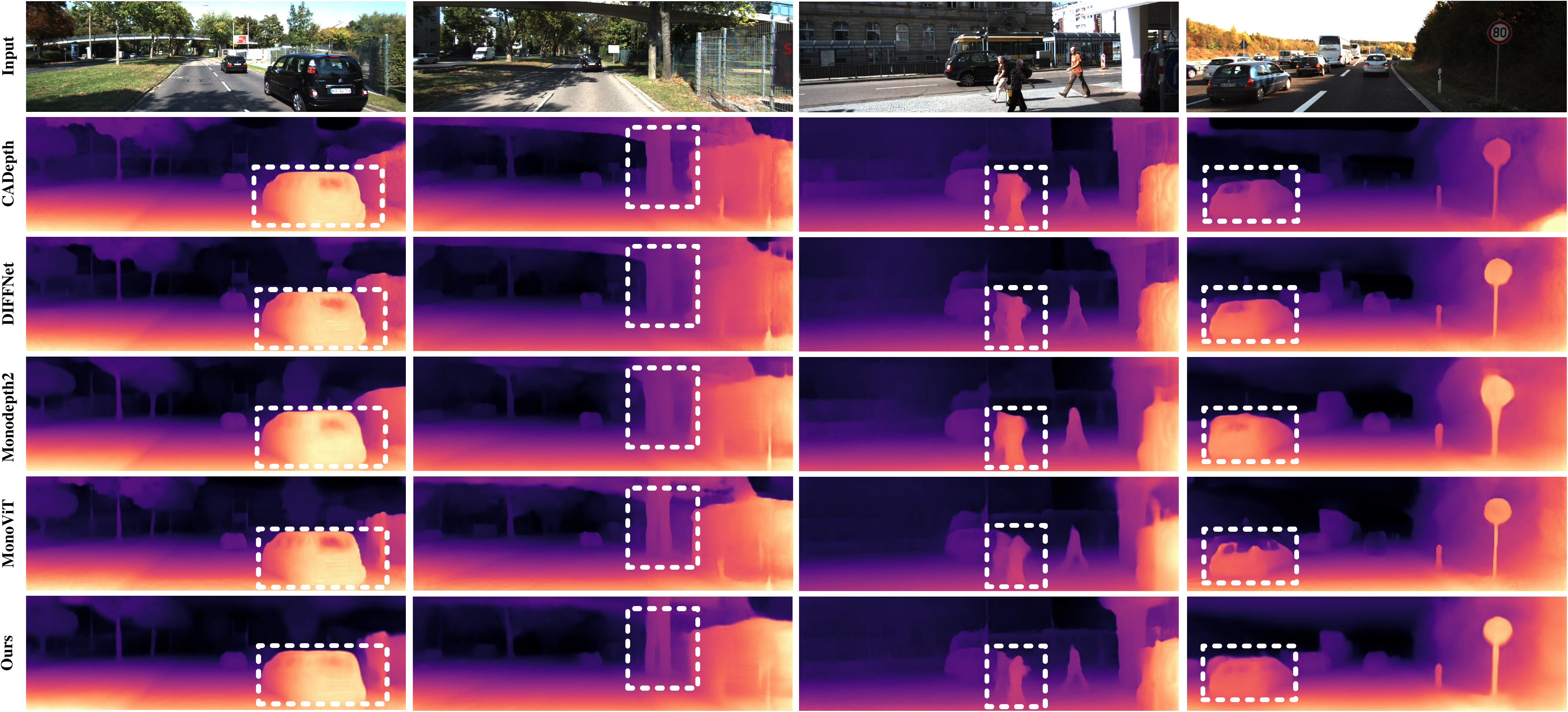}
\caption{Qualitative results on the KITTI \cite{kitti}. Our proposed model in the last row produces much superior depth maps than others, which are reflected in the quantitative results in Table \ref{table_1}.}
\label{fig_kitti}
\end{figure*}
\section{Experiments}
\subsection{Implementation Details}
We use PyTorch to implement our model and the backbone includes ResNet18, ResNet50, Swin Transformer-T, MPViT-S, pretrained on the ImageNet1K dataset \cite{deng2009imagenet}. There are a total of 20 training epochs. In the first epoch, $\gamma$ in the $\mathcal{L}_{total}$ item is set to 0, and from the second epoch of training, we load the model saved in the last epoch for teacher-student distillation and the $\mathcal{L}_{total}$ item is set to 0.1. The initial learning rate for the optimizer we employ, AdamW \cite{admaw}, is set to 1e-4. Inputs with a resolution of 640$\times$192 are trained with a single A5000 GPU, and the batch size is set to 12. Inputs with a resolution of 1024$\times$320 adopt distributed data parallel training, requiring 4 GPUs, and the batch size of a single GPU is set to 4.

\subsection{Datasets}

\textbf{KITTI} \cite{kitti}. The KITTI dataset provides 61 scenes from cities, residential areas, roads, and campuses, utilizing a typical image size of 1242$\times$375. We follow the data split of Eigen et al. \cite{eigen} and Zhou et al.'s \cite{zhou} preprocessing. There are 39,810 monocular triplets for training and 4,424 for validation. In the comparative experiment with other methods, we conduct on the test set \cite{eigen} containing 697 images, which provides 652 depth ground-truth labels. We report results using the per-image
median ground-truth scaling~\cite{zhou} during evaluation.

\begin{table*}[h]
    \renewcommand{\arraystretch}{1.2}
    \centering
    \newcolumntype{L}[1]{>{\raggedright\let\newline\\\arraybackslash\hspace{0pt}}m{#1}}
    \newcolumntype{C}[1]{>{\centering\let\newline\\\arraybackslash\hspace{0pt}}m{#1}}
\newcolumntype{R}[1]{>{\raggedleft\let\newline\\\arraybackslash\hspace{0pt}}m{#1}}
    \resizebox{2.0\columnwidth}{!}{
    \begin{tabular}{L{45mm} || C{10.5mm} C{10.5mm} C{10.5mm} C{13mm}| C{13mm} C{13mm} C{13mm}}
        \hline
        Method& Abs Rel & Sq Rel & RMSE & RMSE log & $\delta<1.25$ & $\delta<1.25^2$ & $\delta<1.25^3$\\
        \hline
        Baseline & 0.103 & 0.740 & 4.458 & 0.179 & 0.896 & 0.966 & 0.983 \\
        Baseline+DA & 0.101 & 0.725 & 4.450 & 0.178 & 0.896 & 0.966 & 0.983 \\
        Baseline+SRD & 0.100 & 0.708 & 4.357 & 0.176 & 0.897 & 0.966 & 0.984 \\ 
        Baseline+SRD+DA & \textbf{0.099} & 0.673 & 4.333 & 0.175 & \textbf{0.898} & 0.966 & 0.984 \\
        Baseline+SRD+DA+MVC (\textbf{full}) & \textbf{0.099} & \textbf{0.659} & \textbf{4.314} & \textbf{0.174} & \textbf{0.898} & \textbf{0.967} & \textbf{0.985} \\
        \hline
        Ours (w/o mask)& \textbf{0.099} & 0.673 & 4.333 & 0.175 & \textbf{0.898} & 0.966 & 0.984\\        
        Ours (w/auto mask \cite{mono2})& 0.102 & 0.707 & 4.364 & 0.175& 0.895 & \textbf{0.967} & 0.984\\
        Ours (w/self-discoverd mask \cite{bian2019unsupervised})& 0.102 & 0.667 & 4.356 & 0.175 & 0.892 & 0.966 & \textbf{0.985}\\
        Ours (w/soft mask) & 0.102 & 0.665 & 4.351 & 0.174 & 0.890 & 0.965 & \textbf{0.985} \\
        \textbf{Ours (w/hard mask)} & \textbf{0.099} & \textbf{0.659} & \textbf{4.314} & \textbf{0.174} & \textbf{0.898} & \textbf{0.967} & \textbf{0.985} \\
        \hline
        Baseline (ResNet18) & 0.115 & 0.903 & 4.863 & 0.193 & 0.877 & 0.959 & 0.981 \\
        \textbf{Ours} (ResNet18)& 0.111 & 0.762 & 4.619 & 0.186 & 0.877 & 0.961 & 0.983 \\  
        \hline
        Baseline (ResNet50) & 0.110 & 0.831 & 4.642 & 0.187 & 0.883 & 0.962 & 0.982 \\
        \textbf{Ours} (ResNet50) & 0.106 & 0.718 & 4.520 & 0.180 & 0.886 & 0.964 & 0.983 \\  
        \hline
        Baseline (Swin-T) & 0.109 &	0.814 &	4.636 &	0.185 &	0.888 &	0.963 &	0.982  \\               
        \textbf{Ours} (Swin-T) & 0.105 & 0.686 & 4.493 & 0.178 & 0.885 & 0.964 & \textbf{0.985} \\  
        \hline
        Baseline (MPViT-S) & 0.103 & 0.740 & 4.458 & 0.179 & 0.896 & 0.966 & 0.983  \\ 
        \textbf{Ours} (MPViT-S) & \textbf{0.099} & \textbf{0.659} & \textbf{4.314} & \textbf{0.174} & \textbf{0.898} & \textbf{0.967} & \textbf{0.985} \\
        \hline
    \end{tabular}}
    \vspace{1em} 
    \caption{\textbf{Ablation results}.{\upshape The baseline model is Monodepth2 \cite{mono2}, which is replaced backbone by MPViT-S \cite{mpvit} .DA denotes disparity alignment, SRD denotes self-reference distillation, and MVC denotes mutilview check.} }
    \label{table_ablation}
\end{table*}

\begin{table*}[h]
    \renewcommand{\arraystretch}{1.2}
    \centering
    \newcolumntype{L}[1]{>{\raggedright\let\newline\\\arraybackslash\hspace{0pt}}m{#1}}
    \newcolumntype{C}[1]{>{\centering\let\newline\\\arraybackslash\hspace{0pt}}m{#1}}
\newcolumntype{R}[1]{>{\raggedleft\let\newline\\\arraybackslash\hspace{0pt}}m{#1}}
    \resizebox{2.0\columnwidth}{!}{
    \begin{tabular}{L{20mm}||C{13mm} C{13mm} C{12mm} || C{10mm} C{9mm} C{9mm} C{13mm}| C{13mm} C{13mm} C{13mm}}
        \hline
        Distillation method& Traning time & Params & FLOPs &Abs Rel & Sq Rel & RMSE & RMSE log & $\delta<1.25$ & $\delta<1.25^2$ & $\delta<1.25^3$\\
        \hline
        w/oDistillation & 42.3h & 27.9M & 20.6G & 0.101 & 0.725 & 4.450 & 0.178 & 0.896 & 0.966 & 0.983 \\ 
        \hline
        w/\cite{poggi2020uncertainty} & 74.7h & 55.8M & 34.8G& 0.100 & 0.695 & 4.373& 0.175 & 0.897 & 0.966 & 0.984	\\
        w/\cite{ren2022adaptive} & 75.9h & 54.5M & 38.0G & 0.100 & 0.695 & 4.337 & 0.176 & 0.897 & 0.965 & 0.983 \\
        w/\cite{hr_depth} & 92.5h & 107.6M & 46.8G & 0.100 & 0.675 & \textbf{4.302} & \textbf{0.174} & 0.896 & 0.966 & 0.984 \\     
        \textbf{w/SRD (Ours)} & \textbf{43.4h} & \textbf{50.5M} & \textbf{23.5G} & \textbf{0.099} & \textbf{0.659} & 4.314 & \textbf{0.174} & 0.898 & \textbf{0.967} & \textbf{0.985} \\
        \hline
    \end{tabular}}
    \vspace{1em} 
    \caption{\textbf{Comparison of the results of different distillation methods}. {\upshape 
    \cite{hr_depth,poggi2020uncertainty,ren2022adaptive} is a two-stage distillation, and SRD (self-reference distillation) is a single-stage online distillation. The Params and FLOPs are calculated according to the input through a forward function of model.} } 
    \label{table_11}
\end{table*}

\textbf{Make3D} \cite{make3d}. The Make3D dataset is an outdoor dataset with a scene similar to KITTI with a fixed image size of 1704$\times$2272, containing a training set of 400 image-depth pairs and a test set of 134 image-depth pairs, which is generally used as a generalization test for monocular depth estimation. Following previous preprocessing \cite{mono2,monovit} on a center crop with a 2$\times$1 ratio, we test the performance of different solutions \cite{mono2,monovit,zhou2021self}.

\subsection{Quantitative Evaluation}
In the comparison experiment, we evaluate images with two resolutions 640$\times$192 and 1024$\times$320 on the KITTI \cite{kitti} dataset, adopting the standard metrics (Abs Rel, Sq Rel, RMSE, RMSE log, $\delta _1<1.25$, $\delta _2<1.25^2$, $\delta _3<1.25^3$) proposed in \cite{eigen} and we cap depth to 80 m per standard practice \cite{mono}. $\delta < t$: $\%$ of ${{\bm{{\rm d}}}}$ satisfies $\left( {\max \left( {\frac{{{{\widehat {\bm{{\rm d}}}}}}}{{{{\bm{{\rm d}}}}}},\frac{{{{\bm{{\rm d}}}}}}{{{{\widehat {\bm{{\rm d}}}}}}}} \right) = \delta  < t} \right)$ for $t = 1.25,{1.25^2},{1.25^3}.$ 

\begin{itemize}
\item $Abs Rel = \frac{1}{\left| {\bm{{\rm T}}} \right|}\sum\nolimits_{\widehat {\bm{{\rm d}}} \in {\bm{{\rm T}}}} {\left| {\widehat {\bm{{\rm d}}} - {\bm{{\rm d}}}} \right|} /{\bm{{\rm d}}},$
\item $Sq Rel = \frac{1}{\left| {\bm{{\rm T}}} \right|}{\sum\nolimits_{\widehat {\bm{{\rm d}}} \in {\bm{{\rm T}}}} {\left\| {\widehat {\bm{{\rm d}}} - {\bm{{\rm d}}}} \right\|} ^2}/{\bm{{\rm d}}},$
\item $RMSE = \sqrt {\frac{1}{\left| {\bm{{\rm T}}} \right|} {{\sum\nolimits_{\widehat {\bm{{\rm d}}} \in {\bm{{\rm T}}}} {\left\| {\widehat {\bm{{\rm d}}} - {\bm{{\rm d}}}} \right\|} }^2}},$
\item $RMSE log = \sqrt {\frac{1}{\left| {\bm{{\rm T}}} \right|} {{\sum\nolimits_{\widehat {\bm{{\rm d}}} \in {\bm{{\rm T}}}} {\left\| {{{\log }_{10}}\widehat {\bm{{\rm d}}} - {{\log }_{10}}{\bm{{\rm d}}}} \right\|} }^2}},$ 
\end{itemize}

We compare the results of several variants of our model trained with different types of supervision. In Table \ref{table_1}, Se denotes training with semantic labels, D for depth supervision, D* for auxiliary depth supervision, and M for mono self-supervision. On evaluation metrics, our model (MPViT-S) outperforms most methods (e.g. \cite{eigen,liu,klodt2018supervising,kundu2018adadepth}) when compared to the depth supervision method. Other metrics outperform the DVSO \cite{yang2018deep}, with the exception of the Abs Rel metric, which is slightly lower. According to the experimental findings, our method closes the gap between supervised and self-supervised monocular depth estimation. In comparison with self-supervised depth estimation methods, ours has a considerable improvement on most evaluation metrics compared to our baseline model Monodepth2 \cite{mono2}. Our accuracy is superior to that of the newly proposed approaches, such as MonoFormer \cite{bae2022monoformer}, CADepth \cite{cad}, and DIFFNet \cite{diffnet} in all metrics. We also contrast the most advanced approach currently available, MonoViT \cite{monovit}. Our results on Abs Rel and $\delta_1$ are comparable, but we perform better on other measures, particularly Sq Rel. On a higher resolution of 1024$\times$320, the accuracy of the model has been further improved. Our monocular depth estimation method demonstrates superior performance compared to state-of-the-art approaches, as evidenced by the results presented in Table \ref{table_1}.

We provide a comparison chart of the visualized outcomes, as seen in Fig.\ref{fig_kitti}, to present our results more intuitively. We indicate with the white dotted box where our method performs better than other methods. For example, in the visualization results, it can be found that MonoViT \cite{monovit} does not perform well in the depth estimation of car mirrors, and Monodepth2 \cite{mono2} does not perform well in overlapping pedestrian occlusion areas.

\subsection{Generalization Evaluation}
We conduct generalization experiments on the Make3D dataset \cite{make3d}. Our model is directly used for the test of the make3D test set without any fine-tuning after training in the KITTI dataset \cite{kitti}. When evaluating metrics, we maintain consistent data preprocessing on a center crop of 2$\times$1 ratio with \cite{mono2,monovit,diffnet,cad}.  

In our generalization testing experiments, we conduct experiments on a test set with 134 samples. We compare some supervised methods with state-of-the-art unsupervised methods. In the generalization of Make3D, our method performs better than the supervised method developed by \cite{eigen} and \cite{liu}. Other metrics are marginally inferior when compared to Laina's supervised method \cite{laina2016deeper}, but our method performs better on the Sq Rel evaluation metric. Our method achieves the best results compared to self-supervised monocular depth estimation methods, and our Sq Rel metric shows the greatest improvement, which is reflected not only in the generalized experiments, but also in the ablation experiments. 
\begin{figure}[h]
\centering
\includegraphics[width=0.48\textwidth]{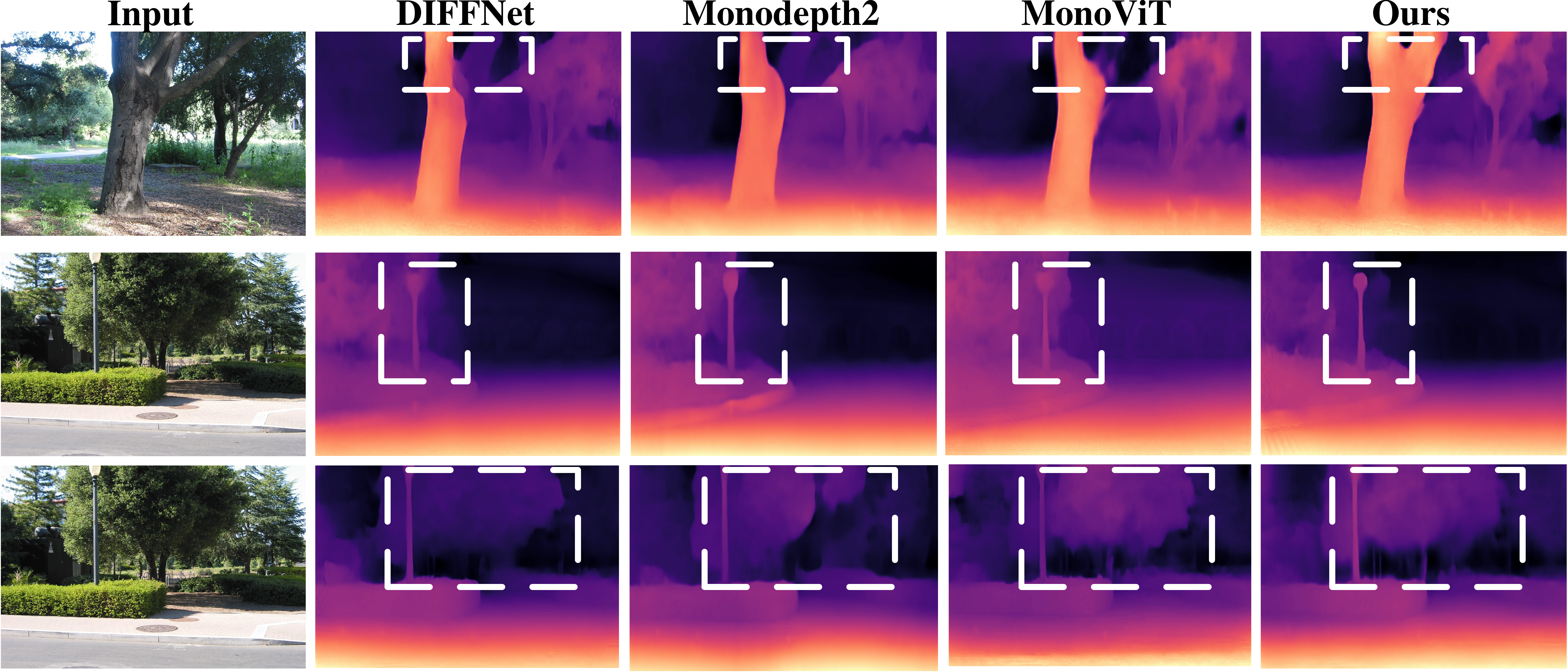}
\caption{Qualitative comparison on the Make3D \cite{make3d} dataset. Predictions by DIFFNet \cite{diffnet}, Monodepth2 \cite{mono2}, MonoViT\cite{monovit} and ours.}
\label{fig_make3d}
\end{figure}
\begin{table}[h]
    \renewcommand{\arraystretch}{1.2}
    \centering
    \newcolumntype{L}[1]{>{\raggedright\let\newline\\\arraybackslash\hspace{0pt}}m{#1}}
    \newcolumntype{C}[1]{>{\centering\let\newline\\\arraybackslash\hspace{0pt}}m{#1}}
\newcolumntype{R}[1]{>{\raggedleft\let\newline\\\arraybackslash\hspace{0pt}}m{#1}}
    \resizebox{1.0\columnwidth}{!}{
    \begin{tabular}{L{20mm} || c c c c}
        \hline
        Method& Abs Rel & Sq Rel & RMSE & $log_{10}$ \\
        \hline
        Eigen\dag \cite{eigen}& 0.428 & 5.079 &  8.389 & 0.149  \\
        Liu\dag \cite{liu}& 0.475 & 6.562 &  10.05 & 0.165 
        \\
        Laina\dag \cite{laina2016deeper}& \textbf{0.204} & \textbf{1.840}&  \textbf{5.683} & \textbf{0.084} 
        \\
        \hline  
        Monodepth \cite{mono}& 0.544 & 10.94 &  11.760 & 0.193 
        \\ 
         Zhou \cite{zhou}& 0.383 & 5.321 &  10.470 & 0.478 
        \\   
          DDVO \cite{wang2018learning}& 0.387 & 4.720 &  8.090 & 0.204 
        \\     
         Monodepth2 \cite{mono2}& 0.322 & 3.589 &  7.417 & 0.163   \\
         CADepth \cite{cad}& 0.319 & 3.564 &  7.152 & 0.158    \\
         HR-Depth \cite{hr_depth}& 0.305 & 2.944 &  6.857 & 0.157 \\ 
         MonoViT \cite{monovit}& 0.286 & 2.758 &  6.623 & 0.147    \\
          \textbf{Ours} &   \textbf{0.252}&   \textbf{1.583}&   \textbf{5.833}&    \textbf{0.114}    \\
        \hline 
    \end{tabular}}
    \vspace{1em}
    \caption{\textbf{Generalization results}. {\upshape The method marked (\dag) denotes the depth supervision method, and the other methods are self-supervised.}}
    \label{table_2}
\end{table}
The results in Table \ref{table_2} demonstrated that, in terms of generalization capacity, our model is much superior to the existing state-of-the-art approaches. Fig.\ref{fig_make3d} displays a depiction of the generalization visualization. We mark the areas worthy of attention with white dotted boxes, and it can also be seen from the visualizations that our method outperforms the others.

\subsection{Ablation Study}
To evaluate the effectiveness of each module of the model, we conduct ablation experiments on the KITTI \cite{kitti} dataset. Our benchmark model is Monodepth2 \cite{mono2}, but for a fair experimental comparison, we replace the backbone network of Monodepth2 from ResNet \cite{resnet} with MPViT-S \cite{mpvit}. Based on the benchmark model, we conduct ablation experiments on three modules, namely, self-reference distillation (SRD), multiview check (MVC) and disparity alignment (DA). The experimental results confirm the effectiveness of the three modules we proposed. From the results in Table \ref{table_ablation}, it can be found that the self-reference distillation has a significant improvement in the Sq Rel metric. 

To verify the effectiveness of the proposed mask (hard mask), we compare against other masking strategies, including the proposed mask (soft mask), the self-discovered mask \cite{bian2019unsupervised} and the auto mask \cite{mono2}. The soft mask and the discovered mask are similar in form, but the difference is that the soft mask is used to re-weight depth supervision loss in distillation and the discovered mask is used to re-weight photometric loss in self-supervised learning. From the experimental results, the strategy of re-weighting the loss improves the Sq Rel metric, but other metrics decrease. The auto mask and the proposed mask (hard mask) both use a binarized mask. The former sets mask to only include the loss of pixels where the reprojection error of the warped image is lower than of the original, unwrapped source image, which is used to remove the motion information in the self-supervised training. The latter filters out outliers with large depth errors output by the teacher model during the distillation process. The distribution of these two masks is not consistent. The former is sparsely distributed in the whole image, while the latter is mainly distributed in the border around the image, and a small part is in the middle area of the image. Compared with these three masks, our proposed mask (hard mask) achieves the best performance among all metrics, as shown in the Table \ref{table_ablation}. 

We compare different backbone networks. We use ResNet \cite{resnet}, Swin Transformer \cite{swin_transformer}, and MPViT \cite{mpvit} as the backbone networks to train the model. We selected several models with relatively small and similar numbers of parameters. These models include ResNet18 with 11.7M parameters, ResNet50 with 25 M parameters, Swin Transformer-T with 28 M parameters, and MPViT-S with 22.8 M parameters. The experimental findings in Table \ref{table_ablation} reveal that our method has considerably enhanced most metrics when compared to the baseline model using the same backbone network, and the model using MPViT as the backbone network has achieved the greatest performance outcomes.

In addition, we compare different distillation method \cite{hr_depth,ren2022adaptive,poggi2020uncertainty} to verify the effectiveness of our self-reference distillation. We compare the computational cost and quantitative results of these methods as shown in Table \ref{table_11}. In terms of computational costs, we compare the training time, the parameters and the FLOPs, where the parameters and the FLOPs are calculated according to the input through a forward function of model. \cite{hr_depth,ren2022adaptive,poggi2020uncertainty} are two-stage distillations, which need to complete the training of the teacher model first, and then distill the student model. The entire distillation process requires two stages of training to be completed separately, increasing the training time, while our single-stage online distillation method basically does not increase training time. \cite{hr_depth} builds complex teacher model to distill, so the parameters and FLOPs are large. \cite{ren2022adaptive} uses ensemble strategy but for the fair comparison, it remains consistent with other methods using a single teacher model. The parameters of \cite{poggi2020uncertainty,ren2022adaptive} are comparable to our method  but their FLOPs are still larger than ours. Overall, our method performs better in quantitative results with minimal computational costs.

\section{Conclusion}

In this paper, we propose a novel self-supervised monocular depth estimation method, employing self-reference distillation to provide depth supervision signals for the student model and introducing a multiview check filter to filter outliers in depth maps. In addition, we propose the disparity offset to solve the disparity misalignment problem caused by the upsampling process. Extensive experiments are carried out on two challenging datasets, including the KITTI and Make3D datasets. The experimental results emphasize the effectiveness and strong generalization of our method. 

{\small
	\bibliographystyle{unsrt}
	\bibliography{ref}

\begin{thebibliography}{10}

\bibitem{huang2022depth}
Jiancai Huang, Zhaohui Jiang, Weihua Gui, Zunhui Yi, Dong Pan, Ke~Zhou, and
  Chuan Xu.
\newblock Depth estimation from a single image of blast furnace burden surface
  based on edge defocus tracking.
\newblock {\em IEEE Transactions on Circuits and Systems for Video Technology},
  32(9):6044--6057, 2022.

\bibitem{song2021monocular}
Minsoo Song, Seokjae Lim, and Wonjun Kim.
\newblock Monocular depth estimation using laplacian pyramid-based depth
  residuals.
\newblock {\em IEEE transactions on circuits and systems for video technology},
  31(11):4381--4393, 2021.

\bibitem{adabins}
Shariq~Farooq Bhat, Ibraheem Alhashim, and Peter Wonka.
\newblock Adabins: Depth estimation using adaptive bins.
\newblock In {\em Proceedings of the IEEE/CVF Conference on Computer Vision and
  Pattern Recognition}, pages 4009--4018, 2021.

\bibitem{newcrfs}
Weihao Yuan, Xiaodong Gu, Zuozhuo Dai, Siyu Zhu, and Ping Tan.
\newblock New crfs: Neural window fully-connected crfs for monocular depth
  estimation.
\newblock {\em arXiv preprint arXiv:2203.01502}, 2022.

\bibitem{meng2021cornet}
Xuyang Meng, Chunxiao Fan, Yue Ming, and Hui Yu.
\newblock Cornet: Context-based ordinal regression network for monocular depth
  estimation.
\newblock {\em IEEE Transactions on Circuits and Systems for Video Technology},
  32(7):4841--4853, 2022.

\bibitem{mono2}
Cl{\'e}ment Godard, Oisin Mac~Aodha, Michael Firman, and Gabriel~J Brostow.
\newblock Digging into self-supervised monocular depth estimation.
\newblock In {\em Proceedings of the IEEE/CVF International Conference on
  Computer Vision}, pages 3828--3838, 2019.

\bibitem{hr_depth}
Xiaoyang Lyu, Liang Liu, Mengmeng Wang, Xin Kong, Lina Liu, Yong Liu, Xinxin
  Chen, and Yi~Yuan.
\newblock Hr-depth: High resolution self-supervised monocular depth estimation.
\newblock In {\em Proceedings of the AAAI Conference on Artificial
  Intelligence}, volume~35, pages 2294--2301, 2021.

\bibitem{ren2022adaptive}
Weisong Ren, Lijun Wang, Yongri Piao, Miao Zhang, Huchuan Lu, and Ting Liu.
\newblock Adaptive co-teaching for unsupervised monocular depth estimation.
\newblock In {\em European Conference on Computer Vision}, pages 89--105.
  Springer, 2022.

\bibitem{poggi2020uncertainty}
Matteo Poggi, Filippo Aleotti, Fabio Tosi, and Stefano Mattoccia.
\newblock On the uncertainty of self-supervised monocular depth estimation.
\newblock In {\em Proceedings of the IEEE/CVF Conference on Computer Vision and
  Pattern Recognition}, pages 3227--3237, 2020.

\bibitem{fapn}
Shihua Huang, Zhichao Lu, Ran Cheng, and Cheng He.
\newblock Fapn: Feature-aligned pyramid network for dense image prediction.
\newblock In {\em Proceedings of the IEEE/CVF International Conference on
  Computer Vision}, pages 864--873, 2021.

\bibitem{li2020semantic}
Xiangtai Li, Ansheng You, Zhen Zhu, Houlong Zhao, Maoke Yang, Kuiyuan Yang,
  Shaohua Tan, and Yunhai Tong.
\newblock Semantic flow for fast and accurate scene parsing.
\newblock In {\em European Conference on Computer Vision}, pages 775--793.
  Springer, 2020.

\bibitem{han2022brnet}
Wencheng Han, Junbo Yin, Xiaogang Jin, Xiangdong Dai, and Jianbing Shen.
\newblock Brnet: Exploring comprehensive features for monocular depth
  estimation.
\newblock In {\em European Conference on Computer Vision}, pages 586--602.
  Springer, 2022.

\bibitem{resnet}
Kaiming He, Xiangyu Zhang, Shaoqing Ren, and Jian Sun.
\newblock Deep residual learning for image recognition.
\newblock In {\em Proceedings of the IEEE conference on computer vision and
  pattern recognition}, pages 770--778, 2016.

\bibitem{swin_transformer}
Ze~Liu, Yutong Lin, Yue Cao, Han Hu, Yixuan Wei, Zheng Zhang, Stephen Lin, and
  Baining Guo.
\newblock Swin transformer: Hierarchical vision transformer using shifted
  windows.
\newblock In {\em Proceedings of the IEEE/CVF International Conference on
  Computer Vision}, pages 10012--10022, 2021.

\bibitem{saxena2005learning}
Ashutosh Saxena, Sung Chung, and Andrew Ng.
\newblock Learning depth from single monocular images.
\newblock {\em Advances in neural information processing systems}, 18, 2005.

\bibitem{eigen}
David Eigen, Christian Puhrsch, and Rob Fergus.
\newblock Depth map prediction from a single image using a multi-scale deep
  network.
\newblock {\em Advances in neural information processing systems}, 27, 2014.

\bibitem{lee2019big}
Jin~Han Lee, Myung-Kyu Han, Dong~Wook Ko, and Il~Hong Suh.
\newblock From big to small: Multi-scale local planar guidance for monocular
  depth estimation.
\newblock {\em arXiv preprint arXiv:1907.10326}, 2019.

\bibitem{shao2022towards}
Shuwei Shao, Ran Li, Zhongcai Pei, Zhong Liu, Weihai Chen, Wentao Zhu, Xingming
  Wu, and Baochang Zhang.
\newblock Towards comprehensive monocular depth estimation: Multiple heads are
  better than one.
\newblock {\em IEEE Transactions on Multimedia}, 2022.

\bibitem{zhou}
Tinghui Zhou, Matthew Brown, Noah Snavely, and David~G Lowe.
\newblock Unsupervised learning of depth and ego-motion from video.
\newblock In {\em Proceedings of the IEEE conference on computer vision and
  pattern recognition}, pages 1851--1858, 2017.

\bibitem{zou2018df}
Yuliang Zou, Zelun Luo, and Jia-Bin Huang.
\newblock Df-net: Unsupervised joint learning of depth and flow using
  cross-task consistency.
\newblock In {\em Proceedings of the European conference on computer vision
  (ECCV)}, pages 36--53, 2018.

\bibitem{cad}
Jiaxing Yan, Hong Zhao, Penghui Bu, and YuSheng Jin.
\newblock Channel-wise attention-based network for self-supervised monocular
  depth estimation.
\newblock In {\em 2021 International Conference on 3D Vision (3DV)}, pages
  464--473. IEEE, 2021.

\bibitem{monovit}
Chaoqiang Zhao, Youmin Zhang, Matteo Poggi, Fabio Tosi, Xianda Guo, Zheng Zhu,
  Guan Huang, Yang Tang, and Stefano Mattoccia.
\newblock Monovit: Self-supervised monocular depth estimation with a vision
  transformer.
\newblock {\em arXiv preprint arXiv:2208.03543}, 2022.

\bibitem{diffnet}
Hang Zhou, David Greenwood, and Sarah Taylor.
\newblock Self-supervised monocular depth estimation with internal feature
  fusion.
\newblock {\em arXiv preprint arXiv:2110.09482}, 2021.

\bibitem{liu2021self}
Lina Liu, Xibin Song, Mengmeng Wang, Yong Liu, and Liangjun Zhang.
\newblock Self-supervised monocular depth estimation for all day images using
  domain separation.
\newblock In {\em Proceedings of the IEEE/CVF International Conference on
  Computer Vision}, pages 12737--12746, 2021.

\bibitem{9578595}
Michaël Ramamonjisoa, Michael Firman, Jamie Watson, Vincent Lepetit, and
  Daniyar Turmukhambetov.
\newblock Single image depth prediction with wavelet decomposition.
\newblock In {\em 2021 IEEE/CVF Conference on Computer Vision and Pattern
  Recognition (CVPR)}, pages 11084--11093, 2021.

\bibitem{chen2021fixing}
Shu Chen, Zhengdong Pu, Xiang Fan, and Beiji Zou.
\newblock Fixing defect of photometric loss for self-supervised monocular depth
  estimation.
\newblock {\em IEEE Transactions on Circuits and Systems for Video Technology},
  32(3):1328--1338, 2021.

\bibitem{choi2020safenet}
Jaehoon Choi, Dongki Jung, Donghwan Lee, and Changick Kim.
\newblock Safenet: Self-supervised monocular depth estimation with
  semantic-aware feature extraction.
\newblock {\em arXiv preprint arXiv:2010.02893}, 2020.

\bibitem{guizilini20203d}
Vitor Guizilini, Rares Ambrus, Sudeep Pillai, Allan Raventos, and Adrien
  Gaidon.
\newblock 3d packing for self-supervised monocular depth estimation.
\newblock In {\em Proceedings of the IEEE/CVF Conference on Computer Vision and
  Pattern Recognition}, pages 2485--2494, 2020.

\bibitem{gurram2021monocular}
Akhil Gurram, Ahmet~Faruk Tuna, Fengyi Shen, Onay Urfalioglu, and Antonio~M
  L{\'o}pez.
\newblock Monocular depth estimation through virtual-world supervision and
  real-world sfm self-supervision.
\newblock {\em IEEE Transactions on Intelligent Transportation Systems},
  23(8):12738--12751, 2021.

\bibitem{spencer2020defeat}
Jaime Spencer, Richard Bowden, and Simon Hadfield.
\newblock Defeat-net: General monocular depth via simultaneous unsupervised
  representation learning.
\newblock In {\em Proceedings of the IEEE/CVF Conference on Computer Vision and
  Pattern Recognition}, pages 14402--14413, 2020.

\bibitem{ranjan2019competitive}
Anurag Ranjan, Varun Jampani, Lukas Balles, Kihwan Kim, Deqing Sun, Jonas
  Wulff, and Michael~J Black.
\newblock Competitive collaboration: Joint unsupervised learning of depth,
  camera motion, optical flow and motion segmentation.
\newblock In {\em Proceedings of the IEEE/CVF conference on computer vision and
  pattern recognition}, pages 12240--12249, 2019.

\bibitem{yang2018lego}
Zhenheng Yang, Peng Wang, Yang Wang, Wei Xu, and Ram Nevatia.
\newblock Lego: Learning edge with geometry all at once by watching videos.
\newblock In {\em Proceedings of the IEEE conference on computer vision and
  pattern recognition}, pages 225--234, 2018.

\bibitem{yang2017unsupervised}
Zhenheng Yang, Peng Wang, Wei Xu, Liang Zhao, and Ramakant Nevatia.
\newblock Unsupervised learning of geometry with edge-aware depth-normal
  consistency.
\newblock {\em arXiv preprint arXiv:1711.03665}, 2017.

\bibitem{guizilini2020semantically}
Vitor Guizilini, Rui Hou, Jie Li, Rares Ambrus, and Adrien Gaidon.
\newblock Semantically-guided representation learning for self-supervised
  monocular depth.
\newblock {\em arXiv preprint arXiv:2002.12319}, 2020.

\bibitem{klingner2020self}
Marvin Klingner, Jan-Aike Term{\"o}hlen, Jonas Mikolajczyk, and Tim
  Fingscheidt.
\newblock Self-supervised monocular depth estimation: Solving the dynamic
  object problem by semantic guidance.
\newblock In {\em European Conference on Computer Vision}, pages 582--600.
  Springer, 2020.

\bibitem{distill}
Cristian Buciluǎ, Rich Caruana, and Alexandru Niculescu-Mizil.
\newblock Model compression.
\newblock In {\em Proceedings of the 12th ACM SIGKDD international conference
  on Knowledge discovery and data mining}, pages 535--541, 2006.

\bibitem{hinton2015distilling}
Geoffrey Hinton, Oriol Vinyals, Jeff Dean, et~al.
\newblock Distilling the knowledge in a neural network.
\newblock {\em arXiv preprint arXiv:1503.02531}, 2(7), 2015.

\bibitem{li2017learning}
Yuncheng Li, Jianchao Yang, Yale Song, Liangliang Cao, Jiebo Luo, and Li-Jia
  Li.
\newblock Learning from noisy labels with distillation.
\newblock In {\em Proceedings of the IEEE International Conference on Computer
  Vision}, pages 1910--1918, 2017.

\bibitem{chen2017learning}
Guobin Chen, Wongun Choi, Xiang Yu, Tony Han, and Manmohan Chandraker.
\newblock Learning efficient object detection models with knowledge
  distillation.
\newblock {\em Advances in neural information processing systems}, 30, 2017.

\bibitem{gupta2016cross}
Saurabh Gupta, Judy Hoffman, and Jitendra Malik.
\newblock Cross modal distillation for supervision transfer.
\newblock In {\em Proceedings of the IEEE conference on computer vision and
  pattern recognition}, pages 2827--2836, 2016.

\bibitem{furlanello2018born}
Tommaso Furlanello, Zachary Lipton, Michael Tschannen, Laurent Itti, and Anima
  Anandkumar.
\newblock Born again neural networks.
\newblock In {\em International Conference on Machine Learning}, pages
  1607--1616. PMLR, 2018.

\bibitem{xie2020self}
Qizhe Xie, Minh-Thang Luong, Eduard Hovy, and Quoc~V Le.
\newblock Self-training with noisy student improves imagenet classification.
\newblock In {\em Proceedings of the IEEE/CVF conference on computer vision and
  pattern recognition}, pages 10687--10698, 2020.

\bibitem{swin-depth}
Zeyu Cheng, Yi~Zhang, and Chengkai Tang.
\newblock Swin-depth: Using transformers and multi-scale fusion for
  monocular-based depth estimation.
\newblock {\em IEEE Sensors Journal}, 21(23):26912--26920, 2021.

\bibitem{yang2021transformer}
Guanglei Yang, Hao Tang, Mingli Ding, Nicu Sebe, and Elisa Ricci.
\newblock Transformer-based attention networks for continuous pixel-wise
  prediction.
\newblock In {\em Proceedings of the IEEE/CVF International Conference on
  Computer Vision}, pages 16269--16279, 2021.

\bibitem{ranftl2021vision}
Ren{\'e} Ranftl, Alexey Bochkovskiy, and Vladlen Koltun.
\newblock Vision transformers for dense prediction.
\newblock In {\em Proceedings of the IEEE/CVF International Conference on
  Computer Vision}, pages 12179--12188, 2021.

\bibitem{wang2021pyramid}
Wenhai Wang, Enze Xie, Xiang Li, Deng-Ping Fan, Kaitao Song, Ding Liang, Tong
  Lu, Ping Luo, and Ling Shao.
\newblock Pyramid vision transformer: A versatile backbone for dense prediction
  without convolutions.
\newblock In {\em Proceedings of the IEEE/CVF International Conference on
  Computer Vision}, pages 568--578, 2021.

\bibitem{mpvit}
Youngwan Lee, Jonghee Kim, Jeffrey Willette, and Sung~Ju Hwang.
\newblock Mpvit: Multi-path vision transformer for dense prediction.
\newblock In {\em Proceedings of the IEEE/CVF Conference on Computer Vision and
  Pattern Recognition}, pages 7287--7296, 2022.

\bibitem{yan2021channel}
Jiaxing Yan, Hong Zhao, Penghui Bu, and YuSheng Jin.
\newblock Channel-wise attention-based network for self-supervised monocular
  depth estimation.
\newblock In {\em 2021 International Conference on 3D Vision (3DV)}, pages
  464--473. IEEE, 2021.

\bibitem{zhou2021self}
Hang Zhou, David Greenwood, and Sarah Taylor.
\newblock Self-supervised monocular depth estimation with internal feature
  fusion.
\newblock {\em arXiv preprint arXiv:2110.09482}, 2021.

\bibitem{alignseg}
Zilong Huang, Yunchao Wei, Xinggang Wang, Wenyu Liu, Thomas~S Huang, and
  Humphrey Shi.
\newblock Alignseg: Feature-aligned segmentation networks.
\newblock {\em IEEE Transactions on Pattern Analysis and Machine Intelligence},
  44(1):550--557, 2021.

\bibitem{shallow}
Adriano Cardace, Pierluigi~Zama Ramirez, Samuele Salti, and Luigi Di~Stefano.
\newblock Shallow features guide unsupervised domain adaptation for semantic
  segmentation at class boundaries.
\newblock In {\em Proceedings of the IEEE/CVF Winter Conference on Applications
  of Computer Vision}, pages 1160--1170, 2022.

\bibitem{mono}
Clément Godard, Oisin~Mac Aodha, and Gabriel~J. Brostow.
\newblock Unsupervised monocular depth estimation with left-right consistency.
\newblock In {\em 2017 IEEE Conference on Computer Vision and Pattern
  Recognition (CVPR)}, pages 6602--6611, 2017.

\bibitem{garg}
Ravi Garg, Vijay~Kumar Bg, Gustavo Carneiro, and Ian Reid.
\newblock Unsupervised cnn for single view depth estimation: Geometry to the
  rescue.
\newblock In {\em European conference on computer vision}, pages 740--756.
  Springer, 2016.

\bibitem{ssim}
Zhou Wang, Alan~C Bovik, Hamid~R Sheikh, and Eero~P Simoncelli.
\newblock Image quality assessment: from error visibility to structural
  similarity.
\newblock {\em IEEE transactions on image processing}, 13(4):600--612, 2004.

\bibitem{wang2018learning}
Chaoyang Wang, Jos{\'e}~Miguel Buenaposada, Rui Zhu, and Simon Lucey.
\newblock Learning depth from monocular videos using direct methods.
\newblock In {\em Proceedings of the IEEE conference on computer vision and
  pattern recognition}, pages 2022--2030, 2018.

\bibitem{petrovai2022exploiting}
Andra Petrovai and Sergiu Nedevschi.
\newblock Exploiting pseudo labels in a self-supervised learning framework for
  improved monocular depth estimation.
\newblock In {\em Proceedings of the IEEE/CVF Conference on Computer Vision and
  Pattern Recognition}, pages 1578--1588, 2022.

\bibitem{yao2018mvsnet}
Yao Yao, Zixin Luo, Shiwei Li, Tian Fang, and Long Quan.
\newblock Mvsnet: Depth inference for unstructured multi-view stereo.
\newblock In {\em Proceedings of the European conference on computer vision
  (ECCV)}, pages 767--783, 2018.

\bibitem{liu}
Fayao Liu, Chunhua Shen, Guosheng Lin, and Ian Reid.
\newblock Learning depth from single monocular images using deep convolutional
  neural fields.
\newblock {\em IEEE transactions on pattern analysis and machine intelligence},
  38(10):2024--2039, 2015.

\bibitem{klodt2018supervising}
Maria Klodt and Andrea Vedaldi.
\newblock Supervising the new with the old: learning sfm from sfm.
\newblock In {\em Proceedings of the European Conference on Computer Vision
  (ECCV)}, pages 698--713, 2018.

\bibitem{kundu2018adadepth}
Jogendra~Nath Kundu, Phani~Krishna Uppala, Anuj Pahuja, and R~Venkatesh Babu.
\newblock Adadepth: Unsupervised content congruent adaptation for depth
  estimation.
\newblock In {\em Proceedings of the IEEE conference on computer vision and
  pattern recognition}, pages 2656--2665, 2018.

\bibitem{yang2018deep}
Nan Yang, Rui Wang, Jorg Stuckler, and Daniel Cremers.
\newblock Deep virtual stereo odometry: Leveraging deep depth prediction for
  monocular direct sparse odometry.
\newblock In {\em Proceedings of the European Conference on Computer Vision
  (ECCV)}, pages 817--833, 2018.

\bibitem{luo2018single}
Yue Luo, Jimmy Ren, Mude Lin, Jiahao Pang, Wenxiu Sun, Hongsheng Li, and Liang
  Lin.
\newblock Single view stereo matching.
\newblock In {\em Proceedings of the IEEE Conference on Computer Vision and
  Pattern Recognition}, pages 155--163, 2018.

\bibitem{guo2018learning}
Xiaoyang Guo, Hongsheng Li, Shuai Yi, Jimmy Ren, and Xiaogang Wang.
\newblock Learning monocular depth by distilling cross-domain stereo networks.
\newblock In {\em Proceedings of the European Conference on Computer Vision
  (ECCV)}, pages 484--500, 2018.

\bibitem{r_msfm}
Zhongkai Zhou, Xinnan Fan, Pengfei Shi, and Yuanxue Xin.
\newblock R-msfm: Recurrent multi-scale feature modulation for monocular depth
  estimating.
\newblock In {\em Proceedings of the IEEE/CVF International Conference on
  Computer Vision}, pages 12777--12786, 2021.

\bibitem{zhou2020constant}
Hang Zhou, David Greenwood, Sarah Taylor, and Han Gong.
\newblock Constant velocity constraints for self-supervised monocular depth
  estimation.
\newblock In {\em Proceedings of the 17th ACM SIGGRAPH European Conference on
  Visual Media Production}, pages 1--8, 2020.

\bibitem{johnston2020self}
Adrian Johnston and Gustavo Carneiro.
\newblock Self-supervised monocular trained depth estimation using
  self-attention and discrete disparity volume.
\newblock In {\em Proceedings of the ieee/cvf conference on computer vision and
  pattern recognition}, pages 4756--4765, 2020.

\bibitem{bae2022monoformer}
Jinwoo Bae, Sungho Moon, and Sunghoon Im.
\newblock Monoformer: Towards generalization of self-supervised monocular depth
  estimation with transformers.
\newblock {\em arXiv preprint arXiv:2205.11083}, 2022.

\bibitem{masoumian2021gcndepth}
Armin Masoumian, Hatem~A Rashwan, Saddam Abdulwahab, Julian Cristiano, and
  Domenec Puig.
\newblock Gcndepth: Self-supervised monocular depth estimation based on graph
  convolutional network.
\newblock {\em arXiv preprint arXiv:2112.06782}, 2021.

\bibitem{kitti}
Andreas Geiger, Philip Lenz, Christoph Stiller, and Raquel Urtasun.
\newblock Vision meets robotics: The kitti dataset.
\newblock {\em The International Journal of Robotics Research},
  32(11):1231--1237, 2013.

\bibitem{cordts2016cityscapes}
Marius Cordts, Mohamed Omran, Sebastian Ramos, Timo Rehfeld, Markus Enzweiler,
  Rodrigo Benenson, Uwe Franke, Stefan Roth, and Bernt Schiele.
\newblock The cityscapes dataset for semantic urban scene understanding.
\newblock In {\em Proceedings of the IEEE conference on computer vision and
  pattern recognition}, pages 3213--3223, 2016.

\bibitem{deng2009imagenet}
Jia Deng, Wei Dong, Richard Socher, Li-Jia Li, Kai Li, and Li~Fei-Fei.
\newblock Imagenet: A large-scale hierarchical image database.
\newblock In {\em 2009 IEEE conference on computer vision and pattern
  recognition}, pages 248--255. Ieee, 2009.

\bibitem{admaw}
Ilya Loshchilov and Frank Hutter.
\newblock Decoupled weight decay regularization.
\newblock {\em arXiv preprint arXiv:1711.05101}, 2017.

\bibitem{bian2019unsupervised}
Jiawang Bian, Zhichao Li, Naiyan Wang, Huangying Zhan, Chunhua Shen, Ming-Ming
  Cheng, and Ian Reid.
\newblock Unsupervised scale-consistent depth and ego-motion learning from
  monocular video.
\newblock {\em Advances in neural information processing systems}, 32, 2019.

\bibitem{make3d}
Ashutosh Saxena, Min Sun, and Andrew~Y Ng.
\newblock Make3d: Learning 3d scene structure from a single still image.
\newblock {\em IEEE transactions on pattern analysis and machine intelligence},
  31(5):824--840, 2008.

\bibitem{laina2016deeper}
Iro Laina, Christian Rupprecht, Vasileios Belagiannis, Federico Tombari, and
  Nassir Navab.
\newblock Deeper depth prediction with fully convolutional residual networks.
\newblock In {\em 2016 Fourth international conference on 3D vision (3DV)},
  pages 239--248. IEEE, 2016.

\end{thebibliography}
}

\begin{IEEEbiography}[{\includegraphics[width=1in,height=1.25in,clip,keepaspectratio]{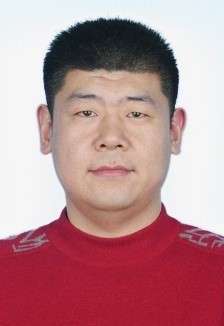}}]{Zhong Liu} received the B.Eng., M.Eng. and Ph.D. degrees from the Harbin Institute of Technology, Harbin,China, in 1991, 1994, and 1997, respectively. He has been with the School of Automation Science and Electronic Engineering, Beihang University, as an Associate Professor from 2000 and as a Professor since 2006. He has published over 50 technical papers in referred journals and conference proceedings and field more than 10 patents. His research interests include bionic CPG mechanism and control, computer vision, parallel mechanism and robotics.
\end{IEEEbiography}

\begin{IEEEbiography}[{\includegraphics[width=1in,height=1.25in,clip,keepaspectratio]{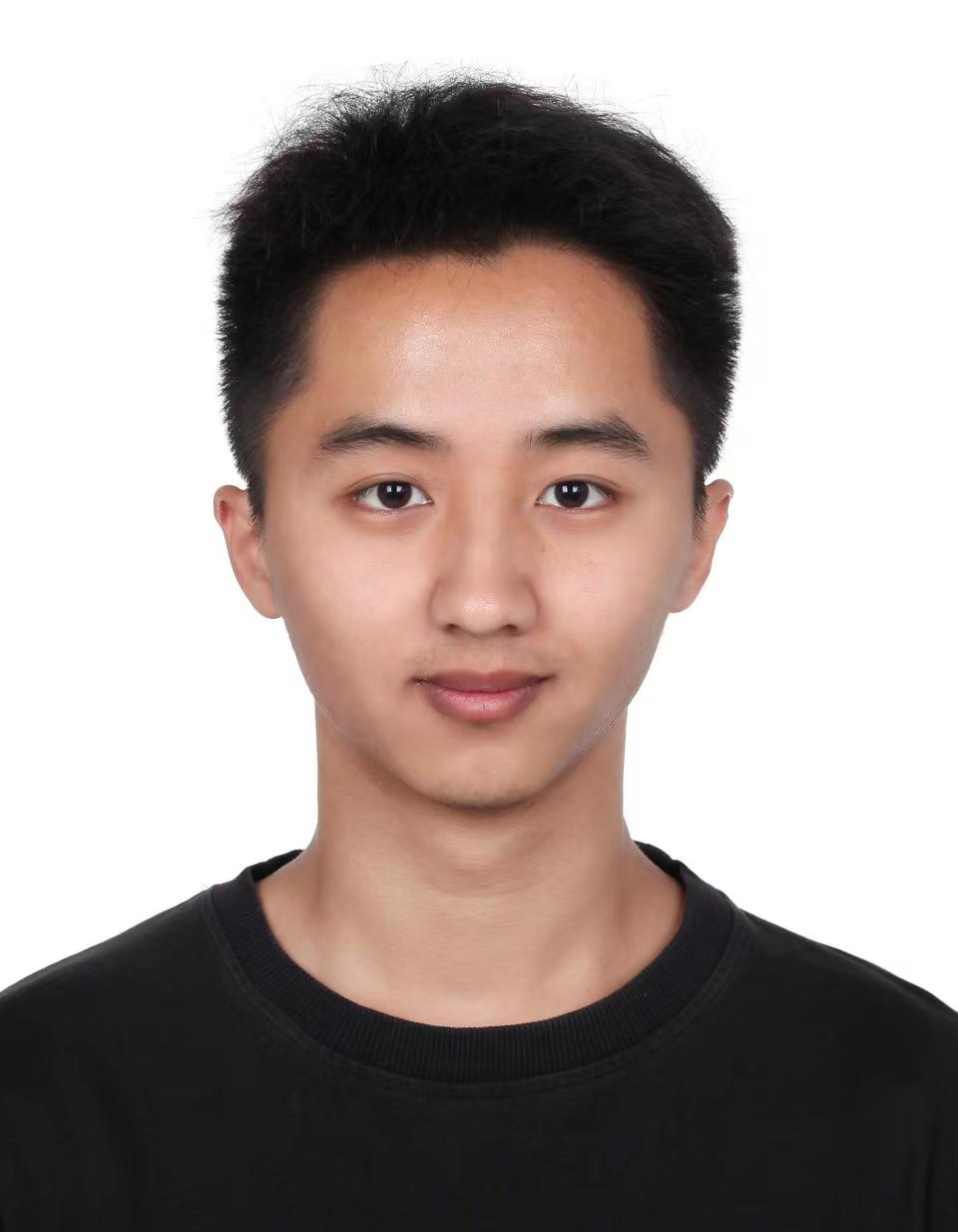}}]{Ran Li} received the bachelor's degree in automation science from BUAA, in 2021, and is currently studying for a master degree with the Beihang University (BUAA), China. His research interests include computer vision and machine learning.
\end{IEEEbiography}

\begin{IEEEbiography}[{\includegraphics[width=1in,height=1.25in,clip,keepaspectratio]{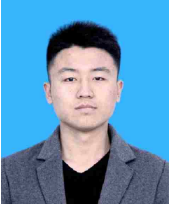}}]{Shuwei Shao} received the B.Eng. from Xidian University, Xi'an, Shanxi, China, in 2019. He is currently pursuing the Ph.D. degree in School of Automation Science and Electrical Engineering, Beihang University, Beijing, China. His current research interests include image registration, depth estimation.
\end{IEEEbiography}

\begin{IEEEbiography}[{\includegraphics[width=1in,height=1.25in,clip,keepaspectratio]{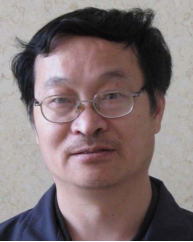}}]{Xingming Wu} received his B.Eng. and M.Eng. degrees from Zhejiang University, in 1981 and 1985, respectively. He is currently an Professor in the School of Automation Science and Electronic Engineering at Beihang University, China. His main research directions are intelligent sensor systems, embedded systems, autonomous mobile robots, and image processing.
\end{IEEEbiography}

\begin{IEEEbiography}[{\includegraphics[width=1in,height=1.25in,clip,keepaspectratio]{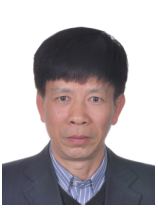}}]{Weihai Chen} received the B.Eng. degree from Zhejiang University, China, in 1982, and the M.Eng. and Ph.D. degrees from Beihang University, China, in 1988 and 1996, respectively. He has been with the School of Automation Science and Electronic Engineering, Beihang University, as an Associate Professor from 1998 and as a Professor since 2007. He has published over 200 technical papers in referred journals and conference proceedings and field more than 20 patents. His research interests include bio-inspired robotics, computer vision, image processing, precision mechanism, automation, and control.
\end{IEEEbiography}

\end{document}